\documentclass[letterpaper, 10 pt, conference]{ieeeconf}  % Comment this line out if you need a4paper

\IEEEoverridecommandlockouts                              % This command is only needed if 
\usepackage{url}
\usepackage{amsmath}

\usepackage[pdftex]{graphicx}
\graphicspath{{../pdf/}{../jpeg/}}
\DeclareGraphicsExtensions{.pdf,.jpeg,.png}

\title{\LARGE \bf
Real-time clustering and multi-target tracking using event-based sensors
}

\author{Francisco Barranco$^{1}$, Cornelia Fermuller$^{2}$ and Eduardo Ros$^{1}$% <-this % stops a space
\thanks{*This work was supported by a Spanish Juan de la Cierva grant (IJCI-2014-21376), partially funded by MINECO-FEDER
TIN2016-81041-R grant, the EU HPB-SGA2 grant (H2020-RIA 785907), the National Science Foundation under grant SMA 1540916, and the NG-UMD seed grant (Object Motion Analysis for Autonomous Systems.}% <-this % stops a space
\thanks{$^{1}$F. Barranco and E. Ros are with the Dept. of Comp. Arch. and Tech., Univ. of Granada, Spain
        {\tt\small fbarranco@ugr.es} and {\tt\small eros@ugr.es}}%
\thanks{$^{2}$C. Fermuller is with the Dept. of Computer Science, UMIACS, University of Maryland, College Park, MD, USA
        {\tt\small fer@umiacs.umd.edu}}%
}

\begin{document}

\maketitle
\thispagestyle{empty}
\pagestyle{empty}

%%%%%%%%%%%%%%%%%%%%%%%%%%%%%%%%%%%%%%%%%%%%%%%%%%%%%%%%%%%%%%%%%%%%%%%%%%%%%%%%
%\begin{abstract}
%Clustering is crucial for many computer vision processing such as robust tracking, object detection and segmentation. This work presents a real-time clustering technique that takes advantage of event-based vision sensors: super high temporal resolution, low latency of a few microseconds, and reduction of computational complexity. Event-based sensors only triggers events when the intensity changes. In such a way, this sensor selects the relevant information reducing redundancy: changes happen only at edges when there is motion in the scene or the camera. Thus, our approach redefines the well-known mean-shift clustering method using asynchronous events instead of conventional frames. Our approach potential is demonstrated with a multi-target tracking application using Kalman filters to smooth the trajectories. We first evaluated our methods using an existing dataset with patterns of different shapes and speeds. Next, the sensor was attached to the Baxter robot in an eye-in-hand setup and analyzed several scenarios with similar patterns and real-world objects in a typical scenario for an action manipulation task. Clustering accuracy achieves an F-measure of 0.95, reducing the computational cost in 88\% compared to the frame-based method. An average error of 2.5 pixels for tracking different objects, achieving a consistent number of clusters along time with similar computational requirements.
%\end{abstract}
\begin{abstract}
Clustering is crucial for many computer vision applications such as robust tracking, object detection and segmentation. This work presents a real-time clustering technique that takes advantage of the unique properties of event-based vision sensors.
%, namely their super high temporal resolution, low latency of a few microseconds, and sparseness of the data.
Since event-based sensors trigger events only when the intensity changes,
the data is sparse, with low redundancy.
%, with most of it originating from edges in the scene.
%in comparison to. In such a way, this sensor selects the relevant information reducing redundancy: changes happen only at edges when there is motion in the scene or the camera. 
Thus, our approach redefines the well-known mean-shift clustering method using asynchronous events instead of conventional frames. The potential of our approach is demonstrated in  a multi-target tracking application using Kalman filters to smooth the trajectories. We  evaluated our method on  an existing dataset with patterns of different shapes and speeds, and a new dataset that we collected. The sensor was attached to the Baxter robot in an eye-in-hand setup monitoring real-world objects in   an action manipulation task. Clustering accuracy achieved an F-measure of 0.95, reducing the computational cost by 88\% compared to the frame-based method. The average error  for tracking was 2.5 pixels and the clustering achieved a consistent number of clusters along time.
\end{abstract}

%%%%%%%%%%%%%%%%%%%%%%%%%%%%%%%%%%%%%%%%%%%%%%%%%%%%%%%%%%%%%%%%%%%%%%%%%%%%%%%%
\section{INTRODUCTION}

%What is clustering, what can be useful for
Clustering aims at partitioning the space creating sets or clusters of elements that are as coherent as possible within the cluster and as different as possible from the elements in the other clusters. It is an unsupervised learning technique for which the cluster assignment or cluster number is unknown. The criterion that determines the classification given a specific element distribution is a distance measure on the space where cluster features are defined.

% Also, say something about tracking
Mean shift clustering is widely used in segmentation and detection \cite{lee_kernel_2013}, tracking and optical flow estimation \cite{zhao_robust_2017}, or feature matching for 3D reconstruction \cite{wei_region_2004}. Visual tracking aims at locating as accurately over time one or more targets (or clusters), in changing scenarios. Real-time visual tracking is a crucial task in Computer Vision, and still a challenge especially with clutter and multiple targets. %The original work on tracking using mean shift \cite{comaniciu_kernel_2003} relies on weighted spatial kernels that use distance functions that are spatially-smooth allowing for the application of gradient optimization techniques. The solution is faster when solving the problem as an optimization, 
%how to solve the problem using conventional cameras, say something about mean shift
Conventional cameras present problems in robotics, especially if real-time processing is required. Cameras capture and transmit a fixed number of frames regardless of the camera motion. The data may have motion blur, or large displacements and occlusions between consecutive frames, causing difficulties for clustering and tracking performance. Even in the absence of motion, when all information is redundant, conventional sensors keep transmitting the very same information resulting in a computational bottleneck.

\begin{figure}[tp]
\begin{center}
\begin{minipage}[b]{0.31\columnwidth}
	\centering
 	\includegraphics[width=\textwidth, height=3.4cm]{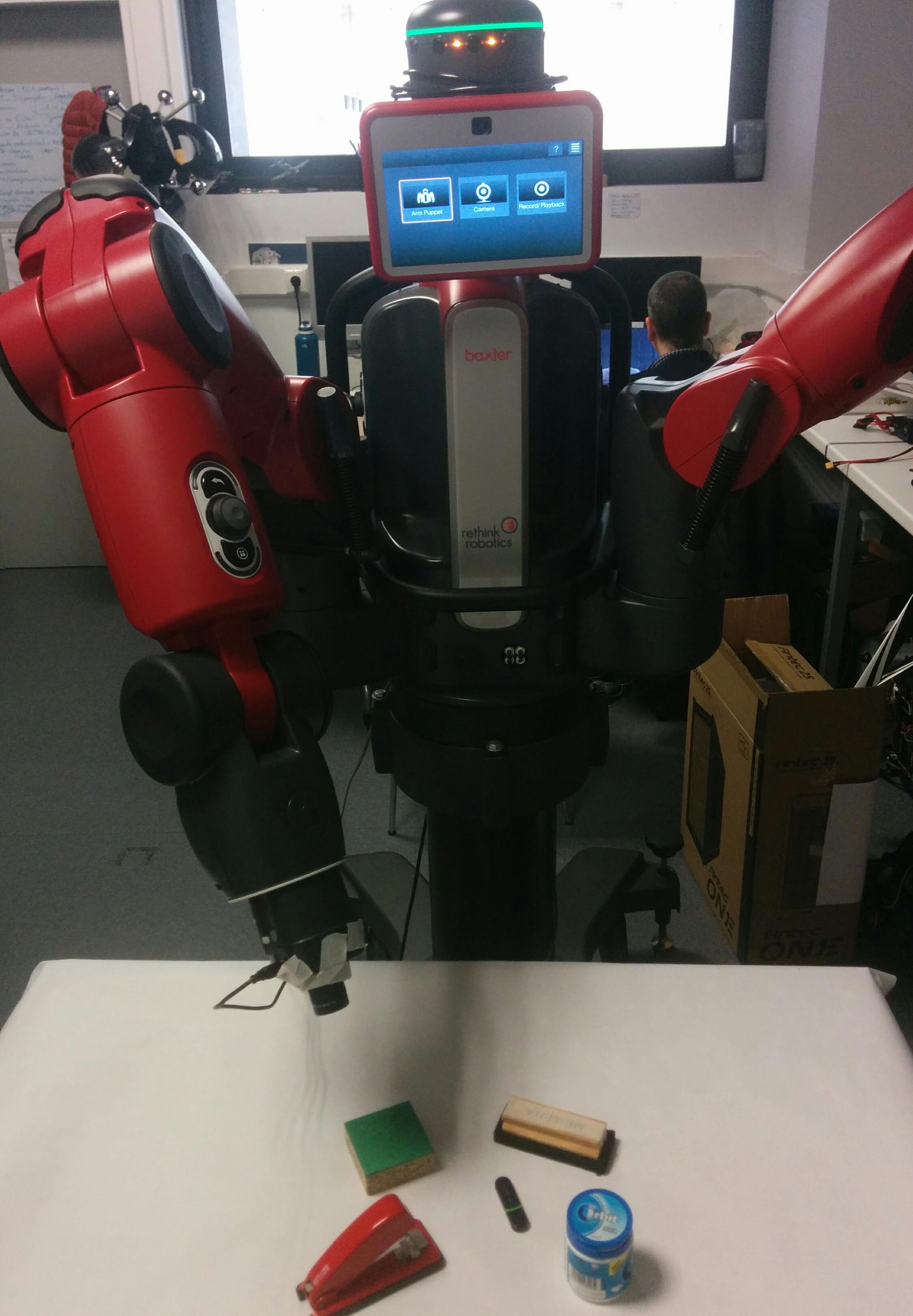}
\end{minipage}
\hspace{0.001cm}
\begin{minipage}[b]{0.31\columnwidth}
	\centering
 	\includegraphics[width=\textwidth, height=3.4cm]{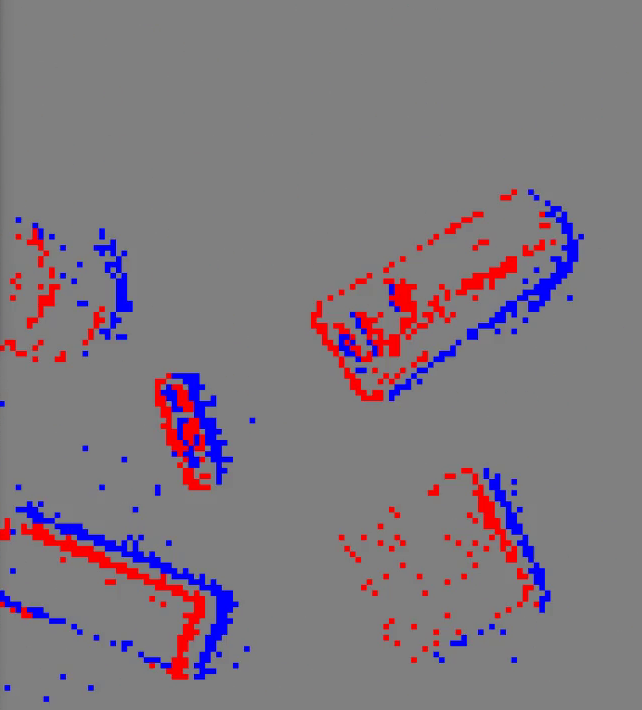}
\end{minipage}
\hspace{0.001cm}
\begin{minipage}[b]{0.32\columnwidth}
	\centering
 	\includegraphics[width=\textwidth, height=3.4cm]{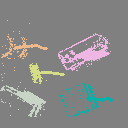}
\end{minipage}
\end{center}
\vspace{-3mm}
\caption{Left to right: A Baxter robot exploring the scenario, event output (positive - red, negative - blue polarity), and clustering and tracking output.}
\label{fig:gen_overview}
\end{figure}
%say something about other methods in the literature for clustering using dvs
We propose a method for real-time event-based clustering and tracking using a Dynamic Vision Sensor (DVS) \cite{lichtsteiner_latency_2008}. These biologically-inspired sensors mimic how eyes perceive motion: getting rid of synchronous frames, these sensors asynchronously transmit events only when pixel intensity changes. These changes are sensed with a very high temporal resolution, and latencies of a few microseconds, orders of magnitude less than in conventional cameras (which have about 33~ms). Perceiving only intensity changes results in the transmission of information only from object edges, reducing data and computational complexity compared to the frame-based methods. A few methods for event-based clustering have been proposed earlier: in \cite{linares_usb3_2015} an FPGA solution is proposed, where clusters are assigned based on the event position and are subsequently tracked; in \cite{mishra_saccade_2017} authors propose the partition of events in \textit{spike groups} to characterize objects for motion segmentation. Additionally, also some methods for event-based tracking have been proposed as in \cite{lagorce_asynchronous_2015}, where authors propose examples of kernels for tracking using oriented Gabor filters and handmade shapes, in \cite{ni_visual_2015} where authors propose a method that handles 2D affine transformations, in \cite{kim_simultaneous_2014} where authors reconstruct the intensity from event information and do the tracking. %, or to track micro-spheres for microscopy in \cite{ni_asynchronous_2012}.  %in \cite{vasco_fast_2016} corners are extracted and tracked. 
%Also, a work that proposed an evaluation metric for event-based tracking \cite{roth_event_2008}. 

We propose a method based on mean-shift clustering that labels each event. Contrarily to previous works, we do not reconstruct intensity or accumulate events to create pseudo-frames. Our real-time solution does not require either providing the number or shape of clusters to be detected and tracked. 
%Finding object contours in the early stages is crucial for further processing such as segmentation, image motion, or object detection.
Although some naive approaches for event-driven computation have been proposed \cite{delbruck_frame_2008}, only the work in \cite{barranco_contour_2015} achieved some event grouping using a supervised method based on Structured Random Forests (not in real-time) and event-based optical flow \cite{barranco_contour_2014}. Clustering events in real-time to label longer contours greatly benefits other applications, besides tracking, such as optical flow or segmentation. Also, a number of methods have developed event-based feature tracking methods \cite{zhu_event_2017}.
%Similarly for tracking, that also benefits from event-based sensors very low latency, specially when objects are moving fast or are occluded. 
Firstly, our clustering is evaluated using a dataset freely available \cite{mueggler_event_2017} that contains sequences with patterns of different shapes. Then, we also apply it in experiments with a humanoid Baxter robot \cite{guizzo_rethink_2012}, and implement a tracking application using Kalman filters to validate it. We mounted our DVS sensor on the Baxter in an eye-in-hand configuration (see Fig.~\ref{fig:gen_overview}). Then, we executed different trajectories at several speeds in an exploratory task of a typical manipulation scenario using real-world objects. All code and data are available\footnote{\url{https://github.com/fbarranco/dvs_meanshift}}.

%comparison with the other works and key selling points of our method, also, explain something about experiments strapping the cameras to the baxter arms

%Include figure of the system running on the Baxter, then the input, and the result with the clustering (real time)

\section{PROPOSED CLUSTERING METHOD}
%Short intro about mean shift, can be more or less technical
%The mean shift clustering method \cite{fukunaga_estimation_1975} is one of the most popular techniques in Computer Vision, lately widely used in filtering and segmentation since Comaniciu and Meer work \cite{comaniciu_mean_2002}. 
Clustering techniques are widely used for analyzing the feature space in problems such as filtering and segmentation. However, most techniques require a-priori knowledge of the number or the shape of clusters and consequently, these techniques are not able to deal with real-world features. The mean shift technique computes for each point the mean of the data distribution (in some multi-dimensional parameter space) in a neighborhood and then, the center of this region is shifted to the computed mean until the processing converges. This algorithm is a nonparametric method that iteratively represents the feature space as a probability density function (PDF), where the modes of the density provide the denser regions and thus, the local maxima of the PDF. Therefore, the solution can be found as an iterative nonparametric density gradient estimation using a specific kernel for estimating the density function. No prior assumption on the number of clusters or their shape is made. Although the original Fukunaga's method \cite{fukunaga_estimation_1975} works iteratively leading to better clustering results, Comaniciu's method \cite{comaniciu_mean_2002} is much faster reducing the number of comparisons in the feature space. We selected a hybrid approach that combines these two, solving the problem as a gradient descent for an optimization \cite{bitsakos_experimental_2010}.  

\subsection{Event-based clustering using mean shift}
%DVS sensor explanation (technical)
%We are not using event integration, how important is not to do that
Event-based sensors only transmit information when the intensity at a specific location changes by a substantial amount; the sensor triggers an event $e(\textbf{x},t,p)$ where $\textbf{x}$ is the location, $t$ the time of the change, and $p$ the polarity (positive or negative intensity change). To exploit the potential of the high temporal timing, we avoid accumulating events; each event is processed asynchronously when it happens. 

%Talk about the decay function for the timestamp (maybe w/ some info from neuros. paper)
The mean-shift solution can be formulated as the gradient descent of the minimization in Eq.~(\ref{eq:meanshift})
\begin{eqnarray}
	 arg\,min_{~\textbf{x}_{i},~p_{i},~f_{\delta}(t_{i})}  \quad \quad \quad \quad \quad \quad \quad \quad \quad \quad \quad \quad \quad \quad \\
	 \sum_{i,j} K_{G}\left(\frac{[\textbf{x}_i,p_i,f_{\delta}(t_{i})] - [\textbf{x}^{0}_{j},p_{j},f_{\delta}(t_{j})]}{h} \right) \nonumber
	\label{eq:meanshift}
\end{eqnarray}
Instead of raw timestamps, we have chosen an exponential decay function defined on the estimated lifetime of events that provides a good dynamic representation of local spatial structures \cite{clady_motion_2017}. In the above equation $f_{\delta}(t) = e^{-\frac{\Delta t}{\tau}}$, where $\tau$ is a parameter that allows tuning the temporal history weight. Thus, we consider 4D data (space, polarity and time), so that the initial values are represented as $[\textbf{x}^{0}_{i}, p^{0}_{i}, f_{\delta}(t^{0}_{i})]$ for pixel $i$. We use a multivariate Gaussian Kernel of the form $K_{G}(\textbf{x})= \frac{1}{\sigma \sqrt{2\pi}} e^{-1/2(\textbf{x}^{T}\textbf{x})}$. Such a kernel although it usually requires more steps to converge, it also results in better clustering. The minimization is defined over a summation of all pairs of pixels $i,j$. The tunnable bandwidth parameter $h$ defines the kernel radius.

In our case, at each iteration the current position is compared to the position of the original set $\textbf{x}^{0}$ as in the classic method \cite{fukunaga_estimation_1975}, but the polarity and time function use the updated values of the previous iteration as done in \cite{comaniciu_mean_2002}. Using this hybrid approach a better clustering is achieved while reducing the computational complexity. Classic mean-shift method is computationally expensive $O(Kn^{2})$ where $n$ is the number of pixels and $K$ the number of iterations before convergence. Instead, we deal only with events that are processed in parallel in small packets of a few hundreds, helping to reduce the computational resource requirements. 

\subsection{Multi-target cluster tracking}
%Kalman filtering - smooth trajectories
Mean-shift clustering has gained popularity in recent years, used as preprocessing in applications such as filtering, segmentation, and tracking. We validate our event-based clustering method for visual tracking of multiple targets. 
Although due to the high temporal resolution the position of every cluster can be tracked pixel-wise, the trajectory of the center of masses is neither accurate nor smooth. For example, after changing the motion direction, different edges may be triggered because events are not triggered for edges that are parallel to the camera motion. Similarly, slow motion may trigger much less events, and thus the calculation of the center of masses may be affected.

Bearing in mind that the additional computation for the tracking must be light in order to keep the clustering processing event-wise, we propose the use of Kalman filters for multi-target tracking. Kalman filters efficiently predict and correct linear processes and are widely used in tracking applications, although usually an accurate system model is required. We assume the reader is comfortable with Kalman filter formulation but more information can be found in \cite{li_multiple_2010}. 

Our Kalman filtering uses a status vector with position $x,y$ and velocity $v_{x},v_{y}$ in pix/s for each center of mass: $\textbf{x}_{k} = [x_{0,k}, y_{0,k}, v_{x}, v_{y}]^T$, where $k$ is the time step. After each step, the goal is to estimate the state vector $\textbf{x}_{k}$ from the measurement vector $\textbf{z}_{k}= [x_{0,k}, y_{0,k}]$ that is only the position of the center of masses provided by clustering. The event-based sensing allows for very accurate small time steps ($\Delta t$) in the update of cluster centers. In our Kalman filter implementation, $A$ is the transition matrix, $H$ the measurement matrix, and we assume a Gaussian cluster-tracking system to define the noise for the process and the measurement.
\[
A=
\begin{bmatrix}
    1 & 0 & \Delta t & 0\\
    0 & 1 & 0 & \Delta t\\
    0 & 0 & 1 & 0\\
    0 & 0 & 0 & 1
\end{bmatrix}
,\quad \quad \quad 
H=
\begin{bmatrix}
    1 & 0 & 0 & 0\\
    0 & 1 & 0 & 0
\end{bmatrix}
\]

%, 
%\[
%H=
%\left[ {\begin{array}{cccc}
%    1 & 0 & 0 & 0\\
%    0 & 1 & 0 & 0\\
%  \end{array} } \right]
%\]
%\label{eq:}
%\end{equation}

One of the advantages of using event sensors is that the timing for measuring the velocity is very accurate thanks to the high temporal resolution; this allows for obtaining very accurate estimates for the Kalman filtering updates. 

\section{EXPERIMENTS}
We configured three sets of experiments to validate our approach for clustering events using mean-shift. Additionally, we also demonstrated the value of our method in a visual multi-target tracking application. For the first experiment, a public dataset \cite{mueggler_event_2017} provides several sequences with different object shapes undergoing various motions in front of an event sensor. In two other experiments we mounted our DVS sensor on a Baxter robot in an eye-in-hand configuration. The first of these experiments was performed moving the Baxter arm in front of a pattern similar to the one of \cite{mueggler_event_2017}. The arm was moved in different directions and at different speeds to analyze our method's sensitivity to orientation and motion direction. In the second set of experiments, we configured a real-world scenario for a manipulation task with the Baxter robot. Different objects were placed on a tabletop, with some occlusions and at different depths from the Baxter. Again, the robot moved its arm in different directions and speeds 
%showing results that do not vary significantly from the first experiments. 
The evaluation of the clustering and tracking accuracy required manually labeling of thousands of chunks of events, randomly chosen and with 4 to 5 clusters in average.  

We use three external error metrics for the evaluation of clustering accuracy: 1) the Adjusted Rand Index (ARI), which measures the similarity between our assignment and the ground-truth ignoring permutations, and it is normalized for chance (random assignments score should be close to 0) to give equal weight to false positives and false negatives; 2) the F-measure (F), which allows weighting false negatives stronger than false positives; 3) the Normalized Mutual Information (NMI), which allows us to measure the quality of our assignment taking into account the number of clusters. For the F-score computation, pairs of events are counted as: if assigned to the same cluster and labeled within the same cluster (TP) or labeled in different clusters (FN); if assigned to different clusters and labeled within the same cluster (FP) or labeled in different clusters (TN).

For the multi-target real-time tracker, the accuracy was evaluated from the distance in the position of the tracked cluster center of masses to the ground-truth labeled cluster center. We also estimate the percentage of valid tracked positions: the proportion of the count of estimates that are estimated with an error lower than a threshold.

We also analyze the time performance for the real-time clustering and the tracking and provide a comparison to the conventional frame-based methods. 

\subsection{Event-based clustering evaluation}
\begin{figure}[tp]
\begin{center}

\begin{minipage}[b]{0.3\columnwidth}
	\centering
 	\includegraphics[width=\textwidth, height=2.5cm]{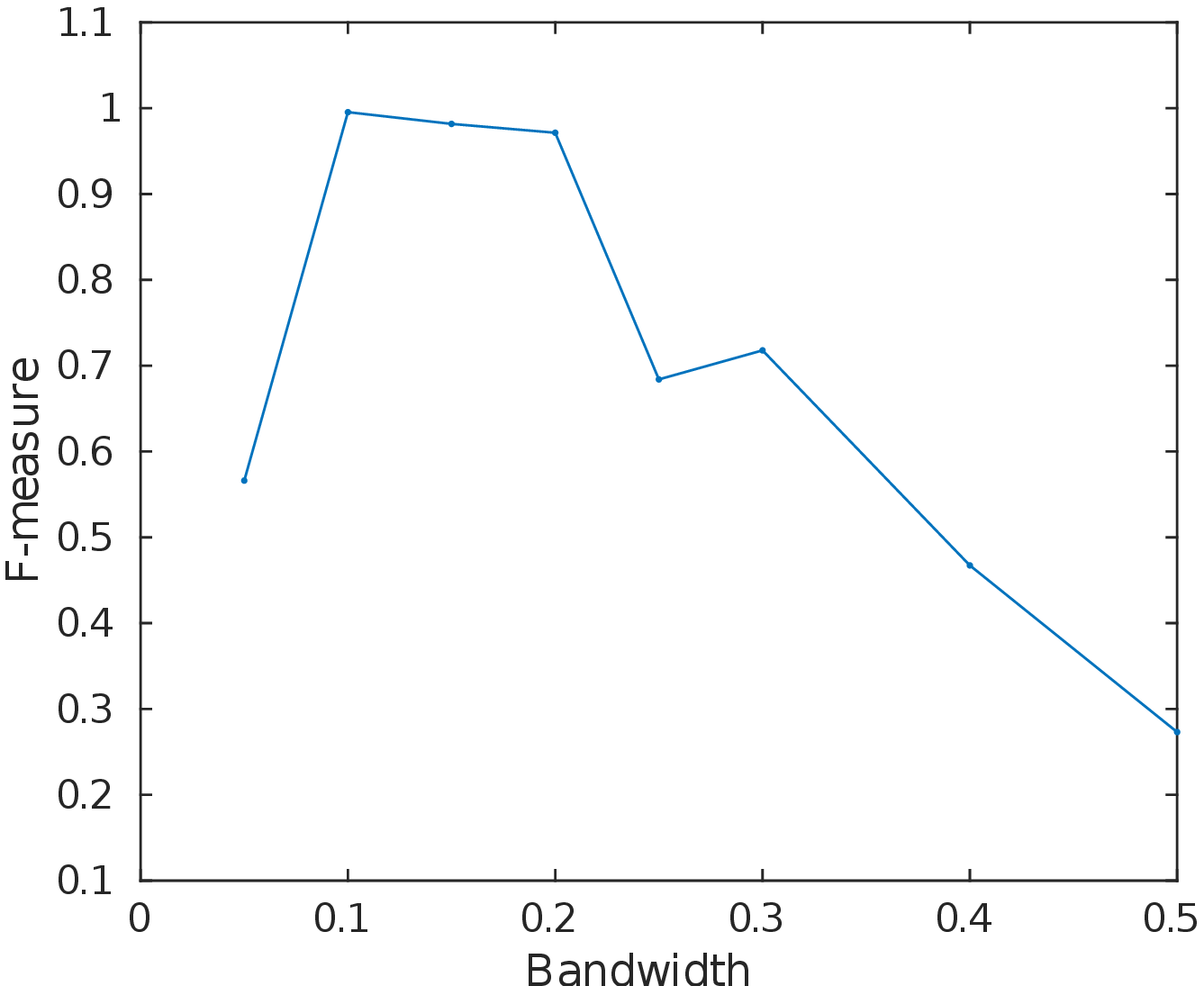}
\end{minipage}
\begin{minipage}[b]{0.34\columnwidth}
	\centering
 	\includegraphics[width=\textwidth, height=2.5cm]{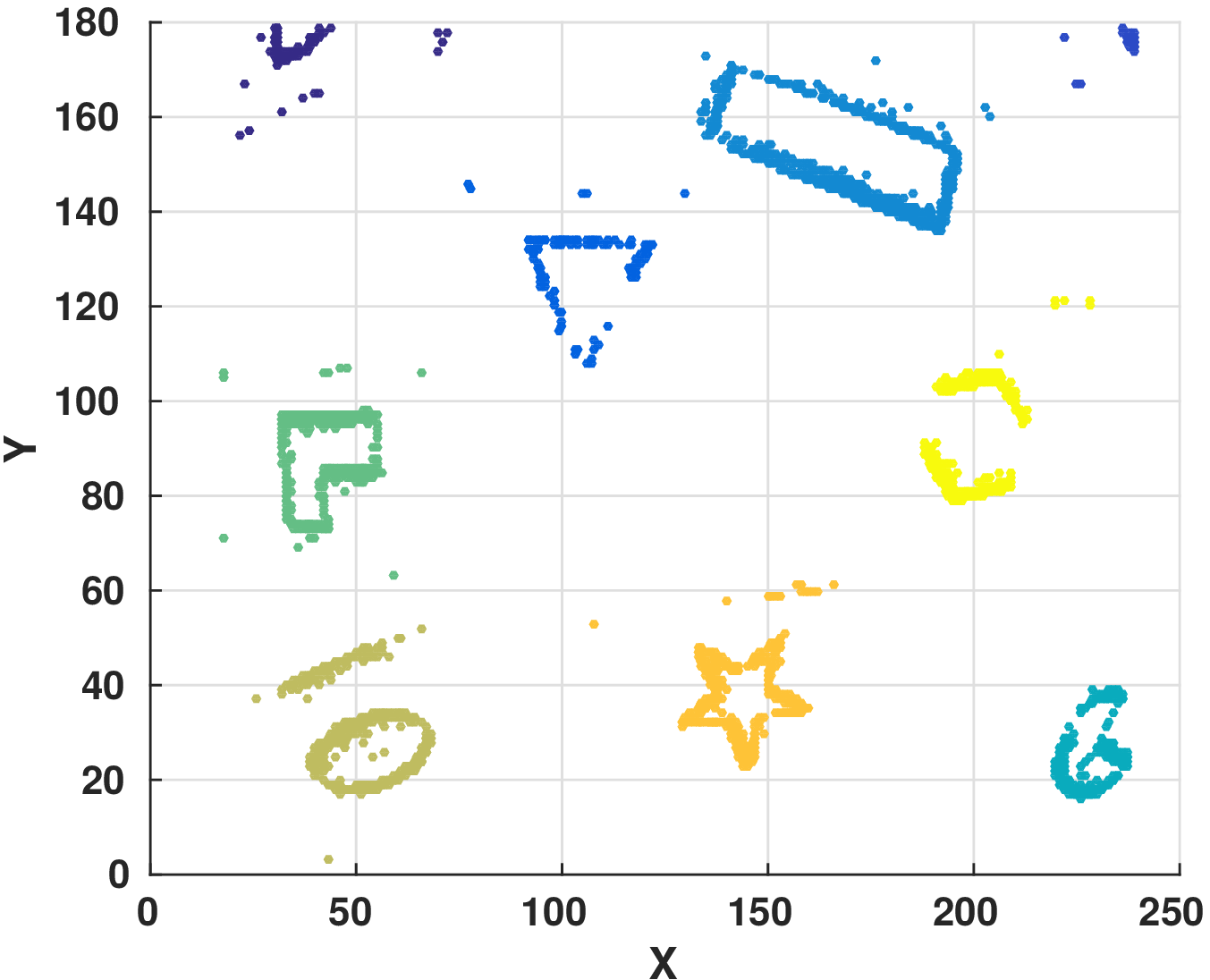}
\end{minipage}
\begin{minipage}[b]{0.33\columnwidth}
	\centering
 	\includegraphics[width=\textwidth, height=2.5cm]{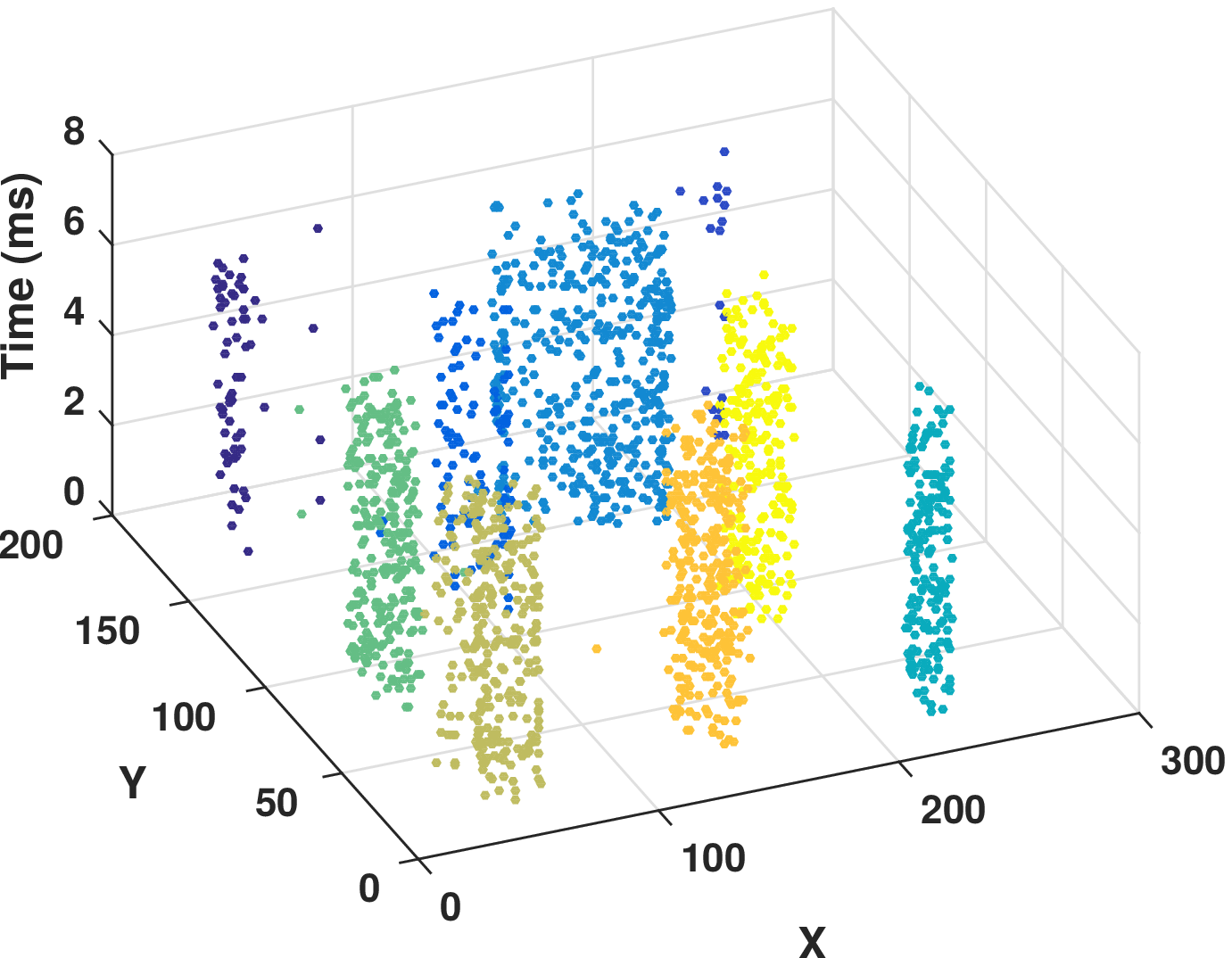}
\end{minipage}

\vspace{0.00cm}

\begin{minipage}[b]{0.3\columnwidth}
	\centering
 	\footnotesize{(a)}
\end{minipage}
\begin{minipage}[b]{0.34\columnwidth}
	\centering
 	\footnotesize{(b)}
\end{minipage}
\begin{minipage}[b]{0.33\columnwidth}
	\centering
 	\footnotesize{(c)}
\end{minipage}

%\begin{minipage}[b]{0.40\columnwidth}
%\centering
%	\begin{minipage}[b]{\columnwidth}
%		\centering
%	 	\includegraphics[width=\textwidth, height=2.4cm]{./figures/cluster3D/cluster3D_1.eps}
%	\end{minipage}
%	\begin{minipage}[b]{\columnwidth}
%		\centering
% 		{\footnotesize  (b)}
%	\end{minipage}
%	%\vspace{0.05cm}
%	\begin{minipage}[b]{\columnwidth}
%		\centering
%	 	\includegraphics[width=\textwidth, height=2.2cm]{./figures/cluster3D/cluster3D_2.eps}
%	\end{minipage}
%\end{minipage}
%
%\begin{minipage}[b]{0.58\columnwidth}
%	\centering
%	{\footnotesize (a)}
%\end{minipage}
%\begin{minipage}[b]{0.40\columnwidth}
%	\centering
%	{\footnotesize (c)}
%\end{minipage}
\end{center}
\vspace{-5mm}
\caption{(a) Evolution of the clustering accuracy given by the F-measure tuning the mean-shift bandwidth parameter $h$. (b-c) Detected clusters in the (x,y,t) space for 5 ms of the \textit{shapes\_rotation} sequence. The cluster labels are encoded with different colors, randomly chosen.}
\label{fig:bandwidth}
\end{figure}

The event-based mean-shift clustering labels each event with the cluster it belongs to. Previous solutions proposed approaches to join event locations from the same contours for problems such as optical flow estimation or segmentation; these solutions create artificial information without solving the problem. Fig.~\ref{fig:bandwidth}b and Fig.~\ref{fig:bandwidth}c show an example of the clustering for a chunk of data from the \textit{shapes\_rotation} sequence. Note the different cluster shapes and sizes and how temporal information enables grouping of events that are triggered almost simultaneously, even if they are not very close. The only parameter for the mean-shift clustering is the so-called bandwidth $h$, which defines the kernel radius. Fig.~\ref{fig:bandwidth}a shows the evolution of the F-measure for clustering with different bandwidth options; in our experiments we use a value of 0.1 that obtains the best accuracy (h is normalized $\in [0,1]$).

%\begin{figure}[tp]
%\begin{center}
%\begin{minipage}[b]{0.25\columnwidth}
%	\centering
% 	\includegraphics[width=\textwidth, height=2.cm]{./figures/motivation/mug2.jpg}
%\end{minipage}
%\begin{minipage}[b]{0.27\columnwidth}
%	\centering
% 	\includegraphics[width=\textwidth, height=2.cm]{./figures/motivation/mug_events.jpg}
%\end{minipage}
%\begin{minipage}[b]{0.27\columnwidth}
%	\centering
% 	\includegraphics[width=\textwidth, height=2.cm]{./figures/motivation/mug_grouping.eps}
%\end{minipage}
%\end{center}
%\vspace{-3mm}
%\caption{Example of previous grouping from \cite{barranco_contour_2014}. Left to right: original image, 10~ms event output (no polarity), long contour grouping.}
%\label{fig:motivation}
%\end{figure}

\begin{table}[h]
	\caption{Clustering accuracy evaluation}
	\label{tab:clustering_evaluation}
\resizebox{\columnwidth}{!}{ 
%	\begin{center}
	\begin{tabular}{|l|l||c|c|c|c|c|}
		\hline
		 & & ARI & NMI & Precision & Recall & Fmeasure\\
		\hline
			\textit{shapes} &	E-MS &  \textbf{0.941} & \textbf{0.952} & \textbf{0.981} & \textbf{0.927} & \textbf{0.951}\\
			\textit{\_translation}				 &	KM   &  0.775 & 0.858 & 0.924 & 0.741 & 0.815\\
		\hline
			\textit{shapes} 	&	E-MS & \textbf{0.947} & \textbf{0.947} & \textbf{0.984} & \textbf{0.929} & \textbf{0.955}\\
			\textit{\_rotation}							&	KM 	 & 0.793 & 0.863 & 0.919 & 0.764 & 0.828\\
		\hline
			\textit{shapes} 		&	E-MS & \textbf{0.912} & \textbf{0.945} & \textbf{0.977} & \textbf{0.888} & \textbf{0.926}\\
			\textit{\_6dof}							&	KM 	 & 0.718 & 0.840 & 0.916 & 0.666 & 0.763\\	
		\hline
			\textit{baxter\_001} 		&	E-MS & \textbf{0.912} & \textbf{0.947} & \textbf{0.901} & \textbf{0.987} & \textbf{0.937}\\
										&	KM 	 & 0.743 & 0.837 & 0.718 & 0.937 & 0.800\\	
		\hline
			\textit{baxter\_002} 		&	E-MS & \textbf{0.949} & \textbf{0.920} & \textbf{0.974} & \textbf{0.958} & \textbf{0.965}\\
										&	KM 	 & 0.694 & 0.750 & 0.700 & 0.870 & 0.769\\
		\hline
			\textit{baxter\_003} 		&	E-MS & \textbf{0.854} & \textbf{0.874} & \textbf{0.872} & \textbf{0.927} & \textbf{0.900}\\
										&	KM 	 & 0.671 & 0.765 & 0.652 & 0.903 & 0.753\\											
		\hline
	\end{tabular}
}
%	\end{center}
\end{table}

\begin{figure*}[tb]
\begin{center}
\begin{minipage}[b]{0.8\textwidth}
	\centering
 	\includegraphics[width=\textwidth, height=2.5cm]{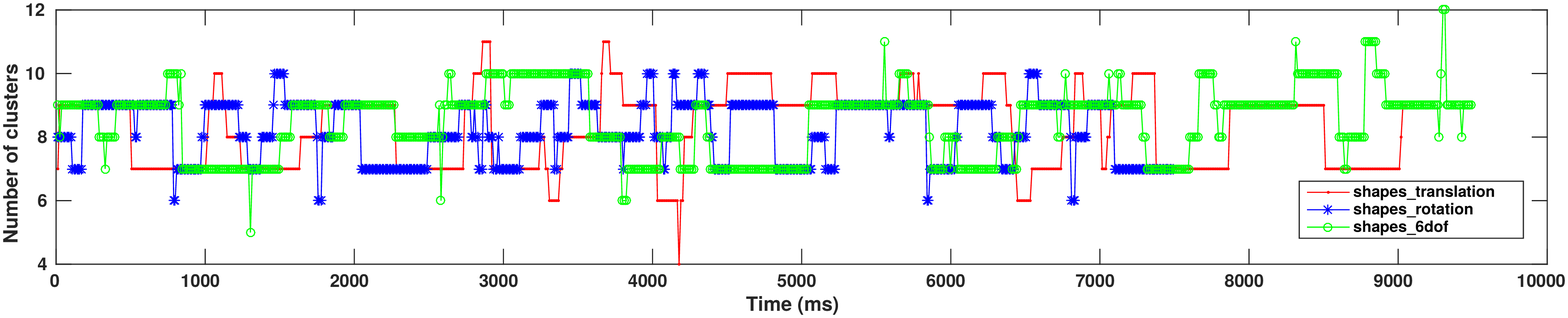}
\end{minipage}
\vspace{0.05cm}
\begin{minipage}[b]{0.8\textwidth}
	\centering
	{\footnotesize  (a)}
\end{minipage}
%\vspace{0.05cm}

\begin{minipage}[b]{0.15\textwidth}
	\centering
 	\includegraphics[width=\textwidth, height=3.4cm]{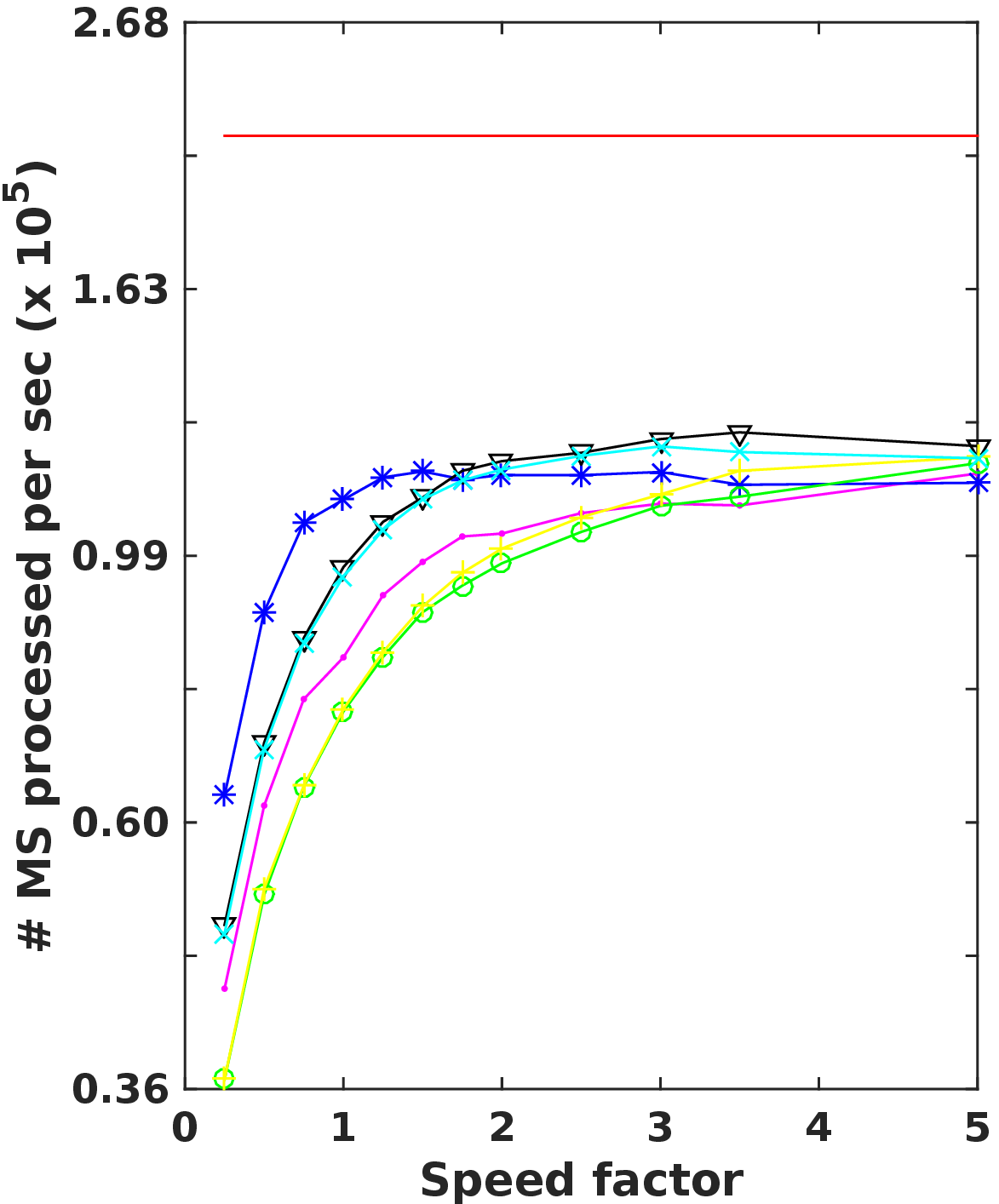}
\end{minipage}
%\hspace{0.05cm}
\begin{minipage}[b]{0.15\textwidth}
	\centering
 	\includegraphics[width=\textwidth, height=3.4cm]{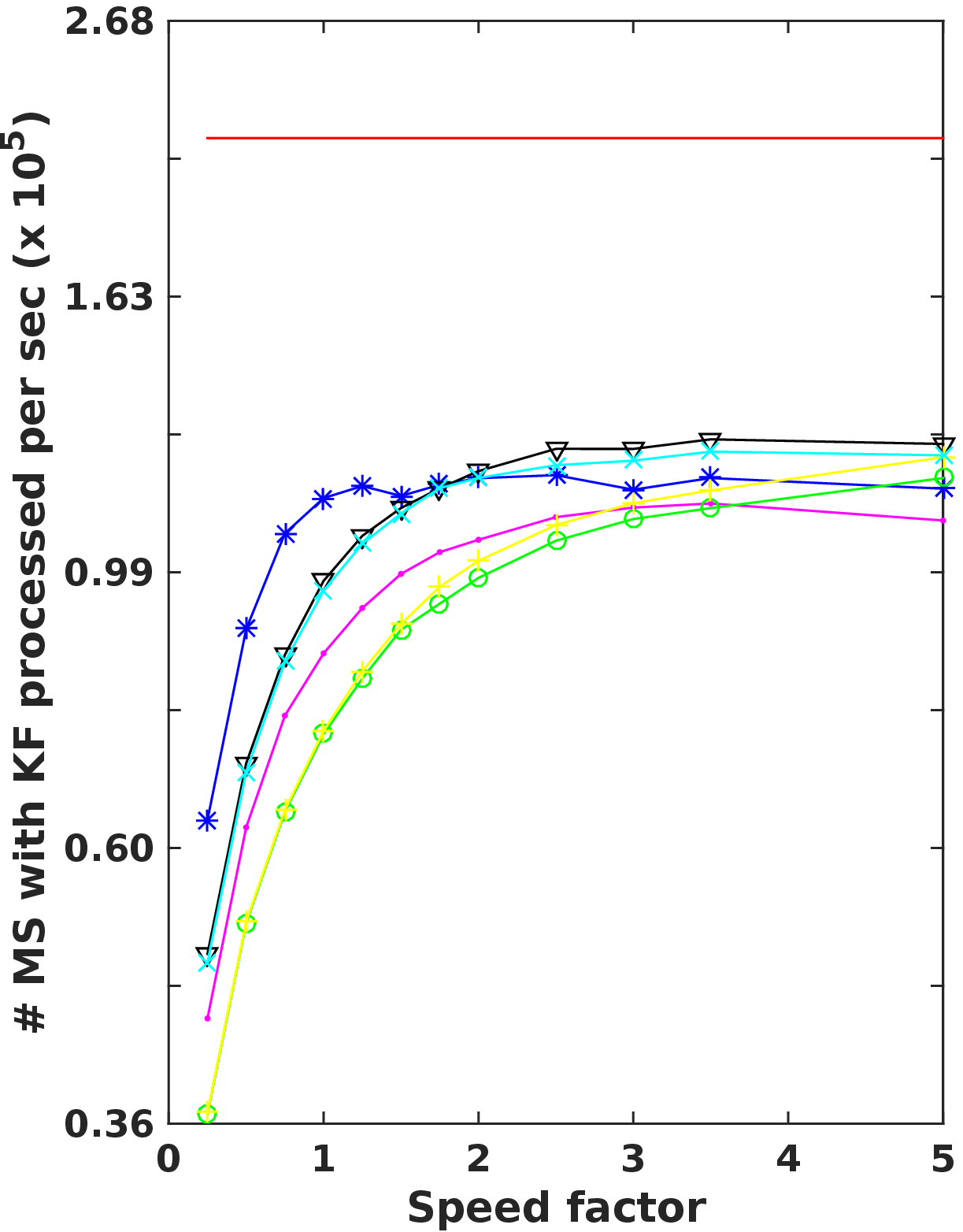}
\end{minipage}
%\hspace{0.05cm}
\begin{minipage}[b]{0.15\textwidth}
	\centering
 	\includegraphics[width=\textwidth, height=3.4cm]{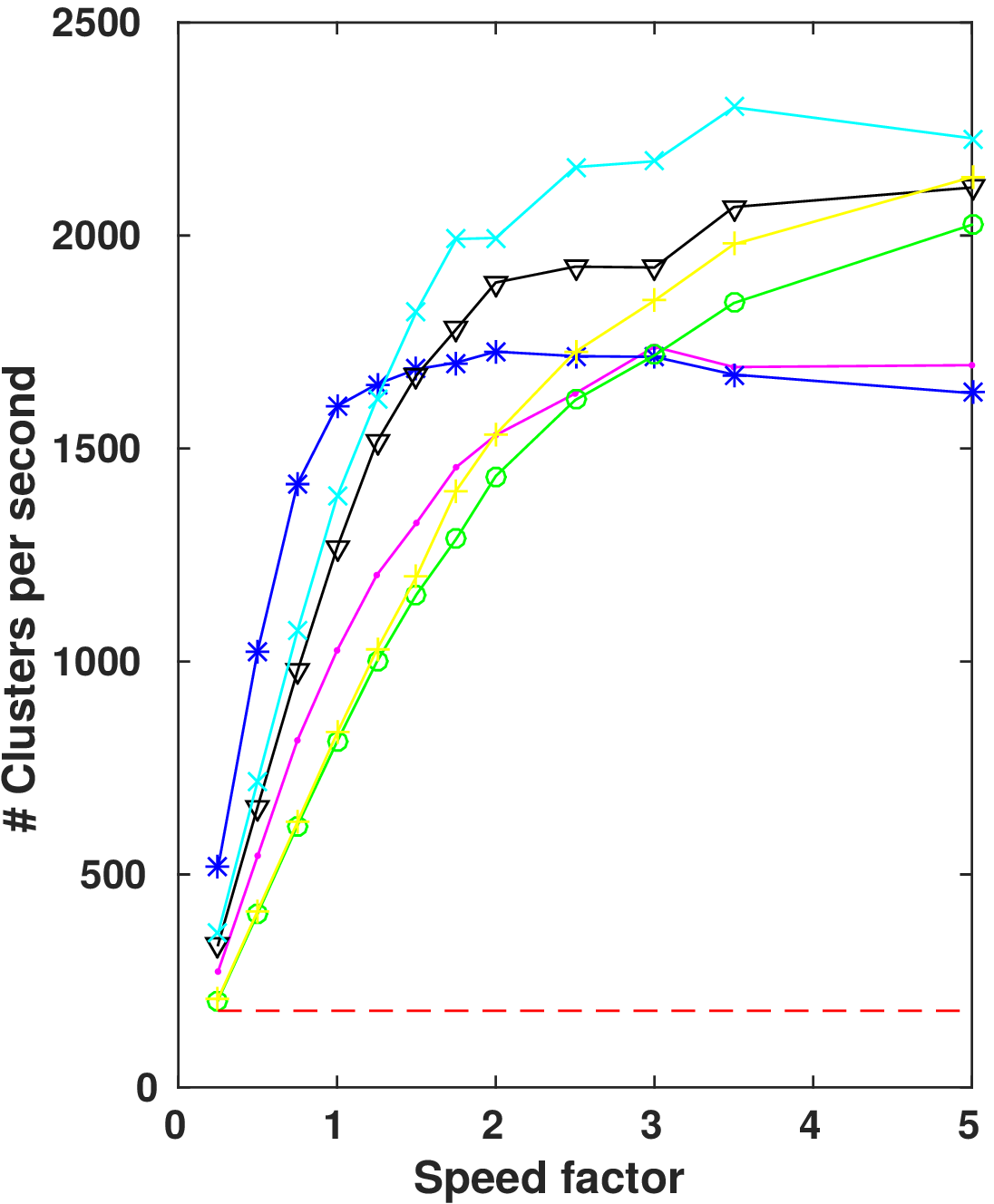}
\end{minipage}
%\hspace{0.05cm}
\begin{minipage}[b]{0.15\textwidth}
	\centering
 	\includegraphics[width=\textwidth, height=3.4cm]{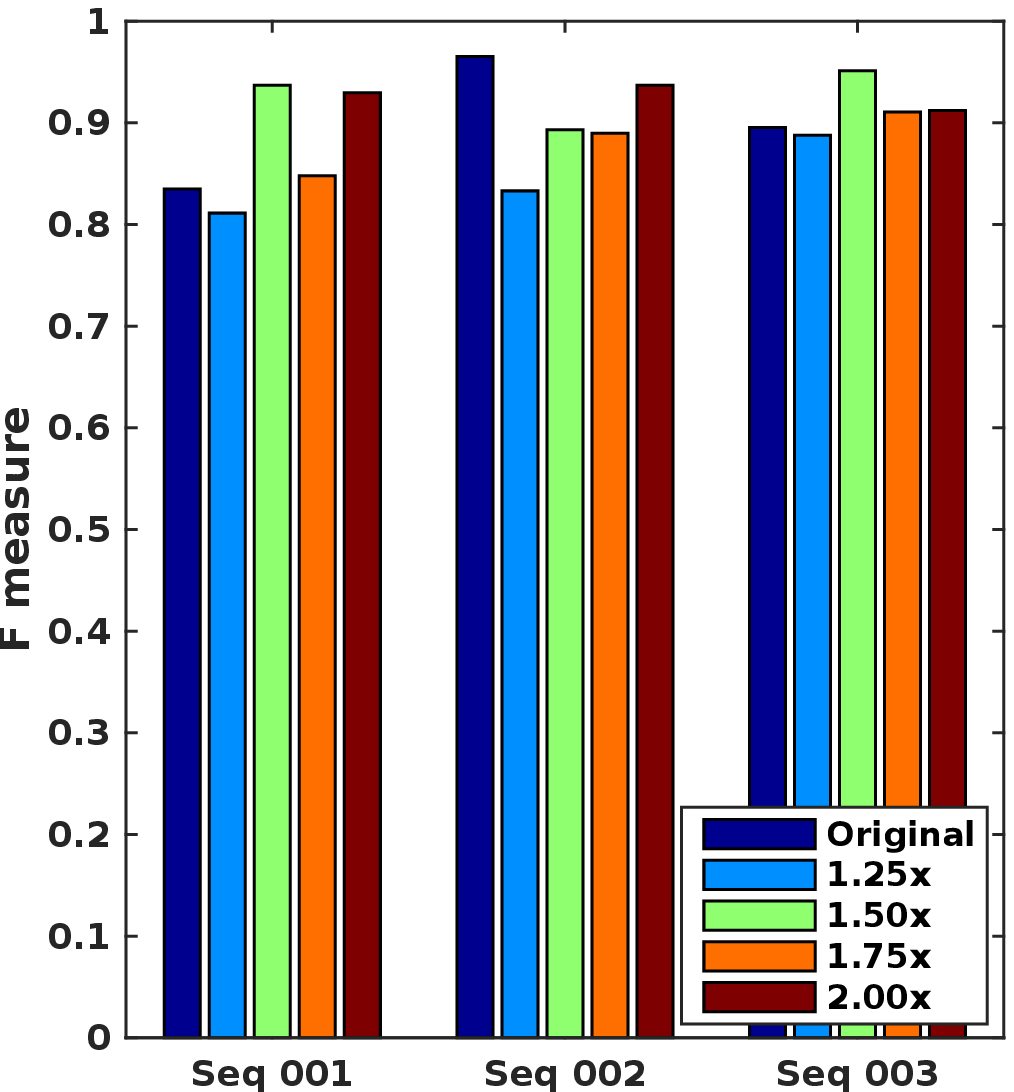}
\end{minipage}
%\vspace{0.05cm}

\begin{minipage}[b]{0.15\textwidth}
	\centering
	{\footnotesize  (b)}
\end{minipage}
%\hspace{0.05cm}
\begin{minipage}[b]{0.15\textwidth}
	\centering
	{\footnotesize  (c)}
\end{minipage}
%\hspace{0.05cm}
\begin{minipage}[b]{0.15\textwidth}
	\centering
 	{\footnotesize  (d)}
\end{minipage}
\begin{minipage}[b]{0.15\textwidth}
	\centering
 	{\footnotesize  (e)}
\end{minipage}
\end{center}
\vspace{-3mm}
\caption{(a) Number of detected clusters of our event-based mean-shift method over time for three sequences; although some objects leave the sensor field of view or get occluded by other objects, the correct number of detected clusters consistently remains. (b) and (c) computational cost represented as the number of mean-shift and mean-shift operations with tracking performed per second (logarithmic scale), when the sensor moves at different speeds: notice almost no difference between the two results and the asymptotic trend for speeds greater than 2.5-3x. The solid red lines represent the computational cost for the frame-based method. (d) Number of clusters detected per second and their growth with higher speed factors. Here the red dashed line at the bottom represents the detection for the frame-based method. (e) F-measure for clustering accuracy with respect to the velocity of the sequence: the first column shows the velocity of the original sequence. No significant differences in the F-measure for higher speeds are shown.}
\label{fig:performance}
\end{figure*}
Before analyzing the results, let us note that we use a preprocessing stage to filter the background activity noise: this filter removes uncorrelated events that is events that do not have any support from past events in the recent time. 

Table~\ref{tab:clustering_evaluation} shows the accuracy table for the event-based clustering method proposed (E-MS) in comparison to the classic K-means clustering method (KM). Due to the lack of event-based clustering methods in the state of the art to compare with, we used as baseline the K-means method. The K-means clustering was performed on the same parameters as our method, namely the position, polarity, and time information. Additionally, the exact number of clusters was also provided to the method, since K-means requires it. We evaluated three sequences from \cite{mueggler_event_2017} and three sequences collected with the Baxter robot in a manipulation task scenario. Our event-based method achieves the best performance for all the metrics and sequences presented, reaching an F-measure that is more precise by approximately 0.13 in the worst case and 0.16 in the best case, and up to 0.2 for the AR Index. Let us remark that the accuracy for \textit{shapes\_6dof} is a bit lower due to the complexity of the scene with shapes rapidly changing with zoom-in zoom-out movements and some shapes are sometimes partially out of the sensor field of view. This results validate our event-based clustering accuracy.

Regarding clustering consistency, Fig.~\ref{fig:performance}a shows the number of detected clusters for three sequences, varying from 7 to 9 during the 10~s shown in the plot. The number of detected clusters remains consistently steady along time, showing no significant differences for different velocities and camera trajectories. Also, the objects in the scene are not visible during the whole sequence due to occlusions or because they leave (partially or completely) the sensor field of view. Fig~\ref{fig:performance}e shows the F-measure value for these sequences and how it varies with the speed. There are no significant variations in the clustering accuracy for higher speeds showing the potential of our method exploiting the high temporal resolution.

We also evaluated the time performance of our event-based computation with respect to the frame-based conventional estimation. The implementation was done under ROS (Robot Operating System): clustering and tracking was implemented in C++ and the Baxter robot uses a Python interface. All code runs in an Intel i7 @ 4GHz with 32 GB of RAM.

For the conventional frame-based mean-shift method, the computational complexity depends on the number of pixels and the number of features; we assume a conventional camera that captures 30 fps (frames per second), with the same spatial resolution than our event-based sensor ($180~\times~240$ or $128~\times~128$ depending on the version). In the frame-based mean-shift method the computation is done for each pixel, resulting in a steady computational rate of $30~\times~180~\times~240 = 1.296 \cdot 10^{6}$ mean-shift processed operations per second. Meanwhile, in our event-based clustering method we are only processing the events that are triggered asynchronously. Therefore, the computational rate in our case depends on the number of events at every time and thus depends on the speed of the camera or the objects in the scene, and the scene structure. The computational complexity in Fig.~\ref{fig:performance}b shows how the mean-shift operations grow with the speed of the sequence, compared to the complexity of the frame-based estimation. The different speeds for the sequences from the dataset in \cite{mueggler_event_2017} are simulated changing the timestamps, while in the robot case the arm moves at different speeds while exploring the real scene. Colors here show the results for different sequences, and the operation rate is computed for 20~s of each sequence. The potential of the event-based method to reduce the computational cost is clear: in the best case, it is reduced in approximately 88\%, and the average computational cost reduction for the original speed is about 83\% (notice the logarithmic scale of the Y axis). 

\begin{figure*}[tb]
\begin{center}
\begin{minipage}[b]{0.32\textwidth}
	\centering
 	\includegraphics[width=\textwidth, height=2.5cm]{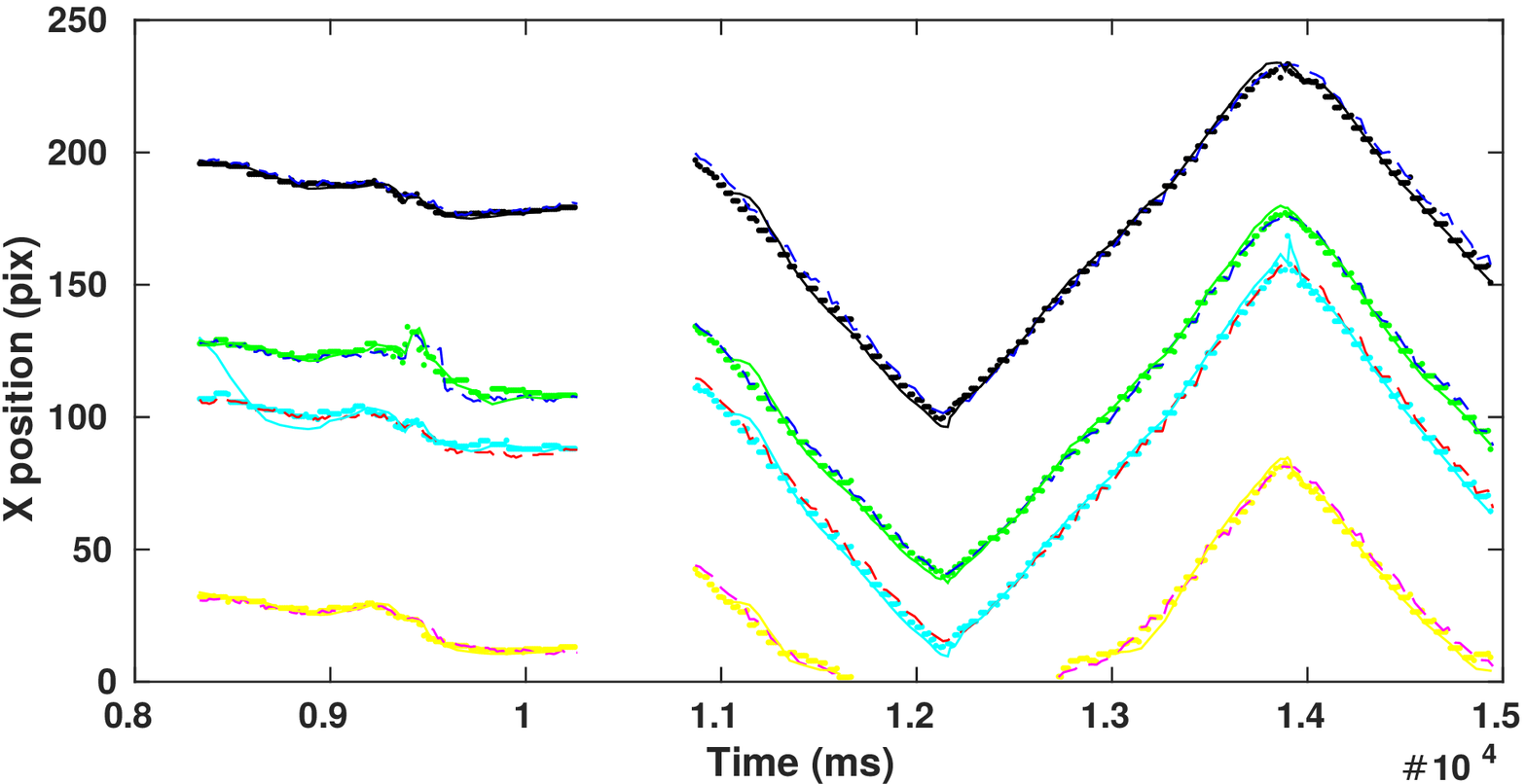}
\end{minipage}
%\hspace{0.05cm}
\begin{minipage}[b]{0.32\textwidth}
	\centering
 	\includegraphics[width=\textwidth, height=2.5cm]{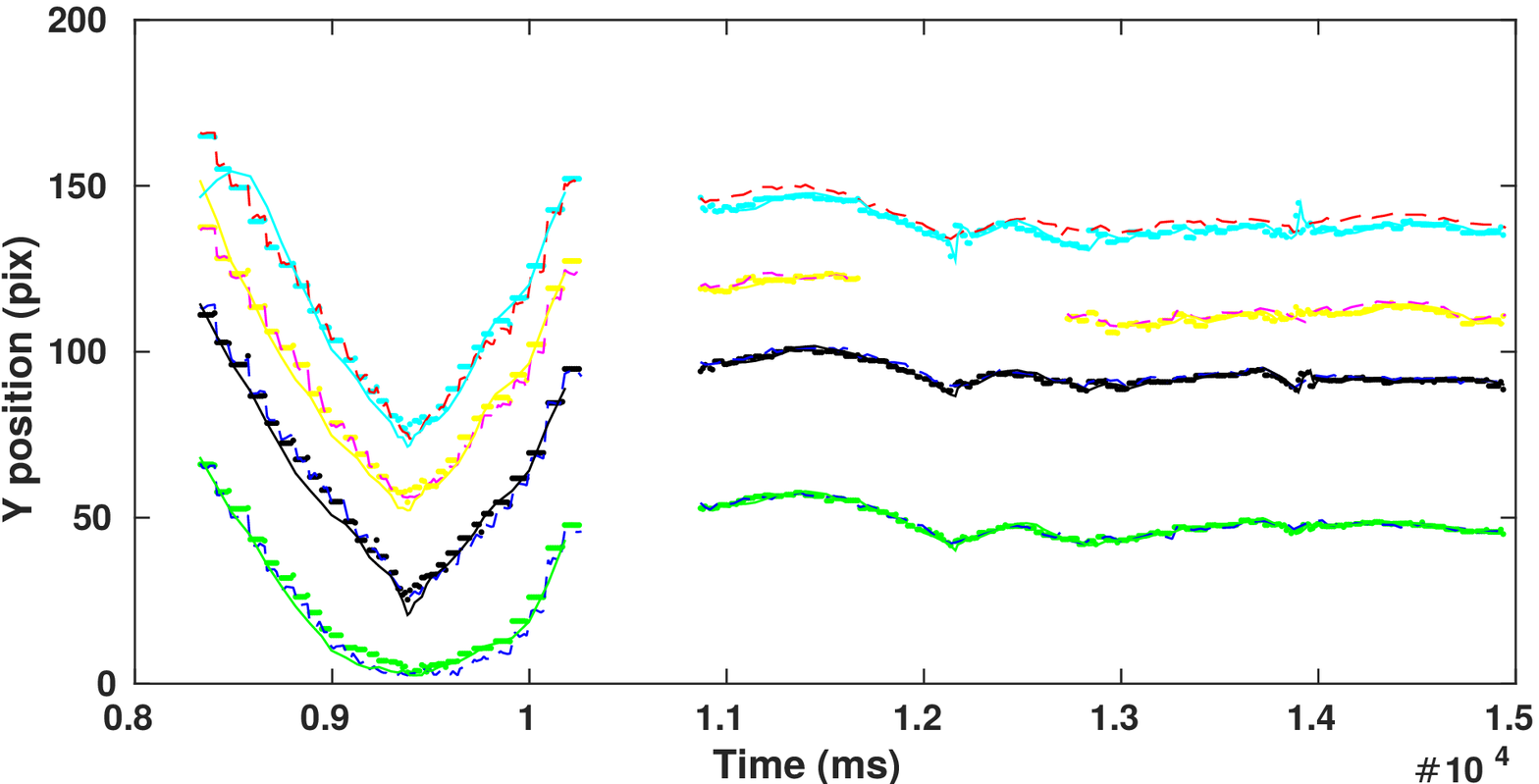}
\end{minipage}
%\hspace{0.05cm}
\begin{minipage}[b]{0.1\textwidth}
	\centering
 	\includegraphics[width=\textwidth, height=2.5cm]{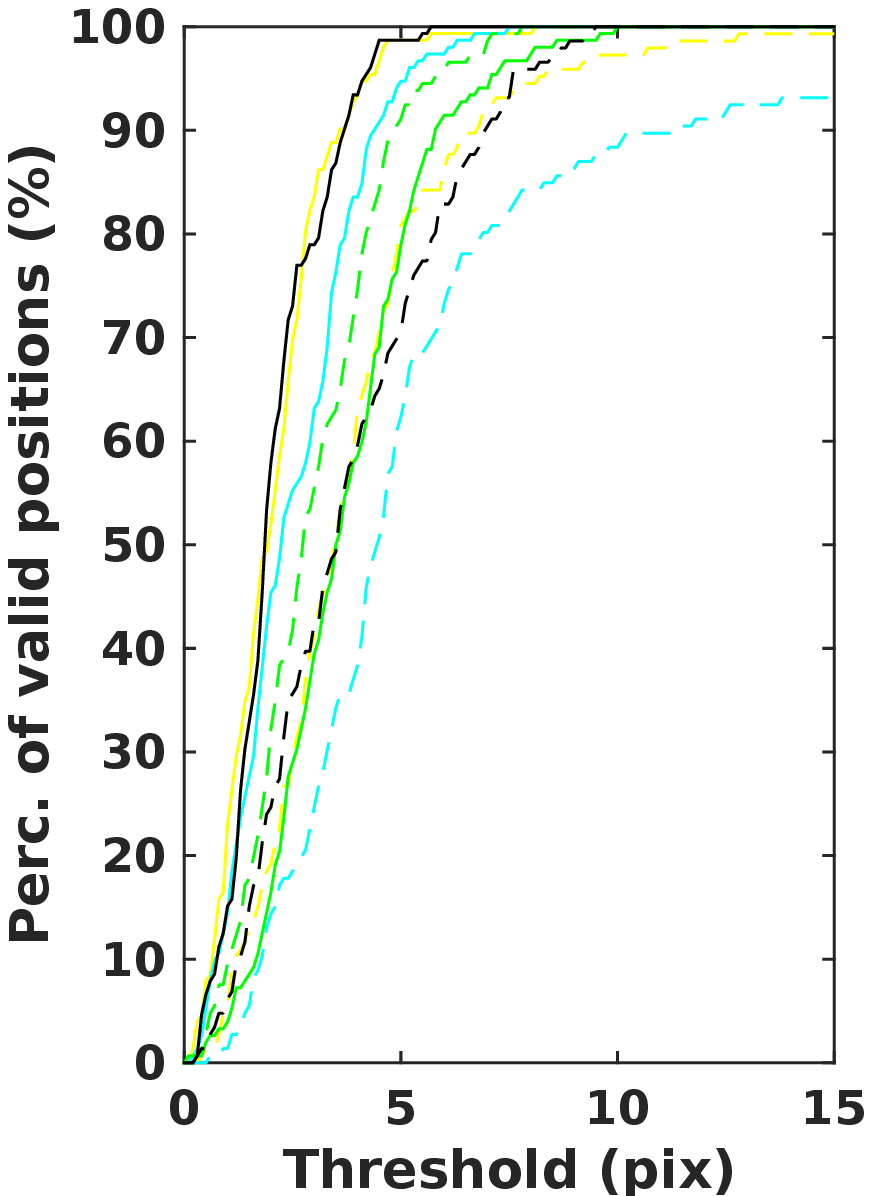}
\end{minipage}
\hspace{2.65cm}
\vspace{0.01cm}

\begin{minipage}[b]{0.32\textwidth}
	\centering
 	\footnotesize{(a)}
\end{minipage}
%\hspace{0.05cm}
\begin{minipage}[b]{0.32\textwidth}
	\centering
	\footnotesize{(d)}
\end{minipage}
%\hspace{0.05cm}
\begin{minipage}[b]{0.1\textwidth}
	\centering
 	\footnotesize{(g)}
\end{minipage}
\hspace{2.65cm}

\vspace{0.01cm}
\begin{minipage}[b]{0.32\textwidth}
	\centering
 	\includegraphics[width=\textwidth, height=2.5cm]{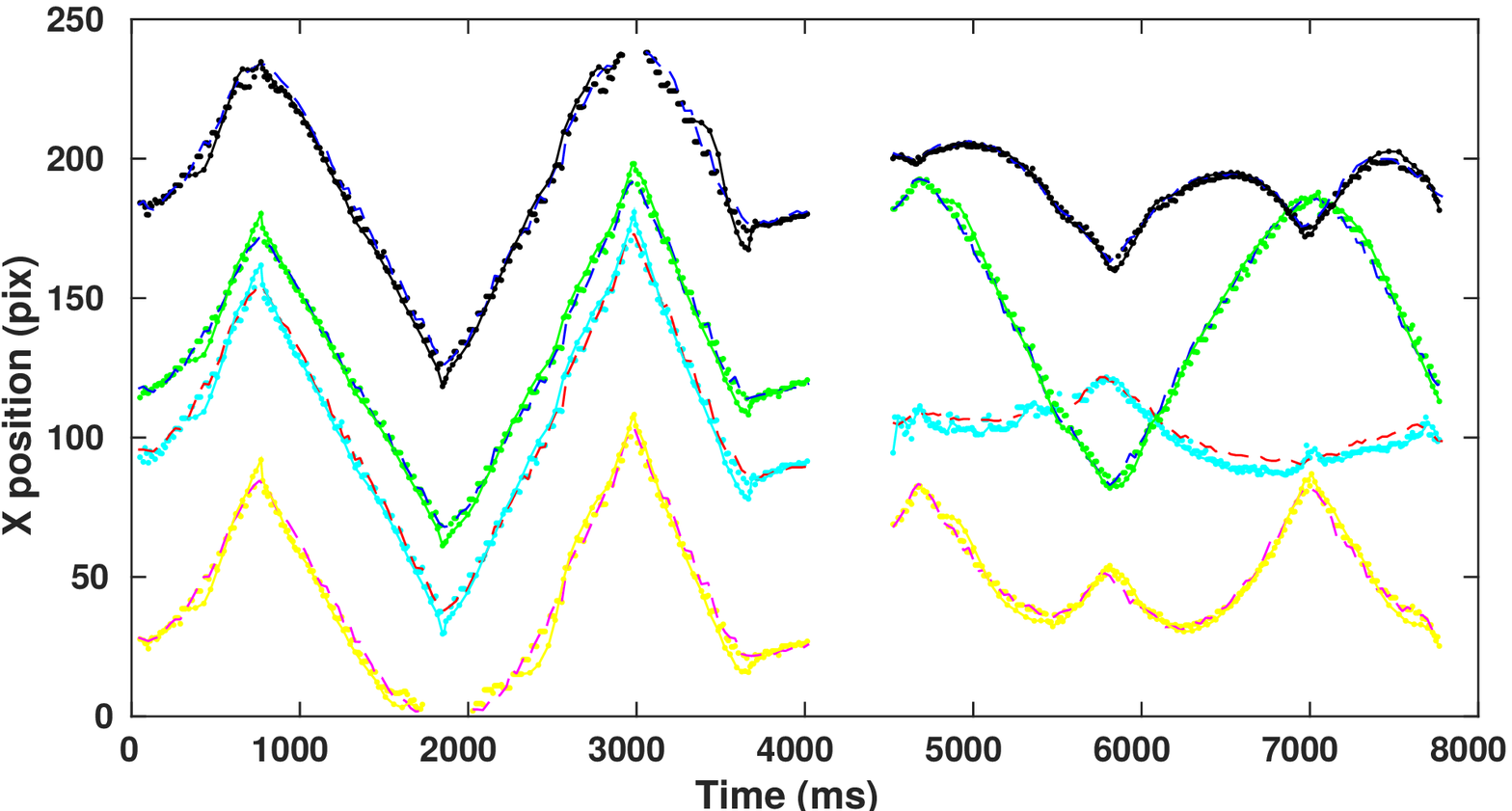}
\end{minipage}
%\hspace{0.05cm}
\begin{minipage}[b]{0.32\textwidth}
	\centering
 	\includegraphics[width=\textwidth, height=2.5cm]{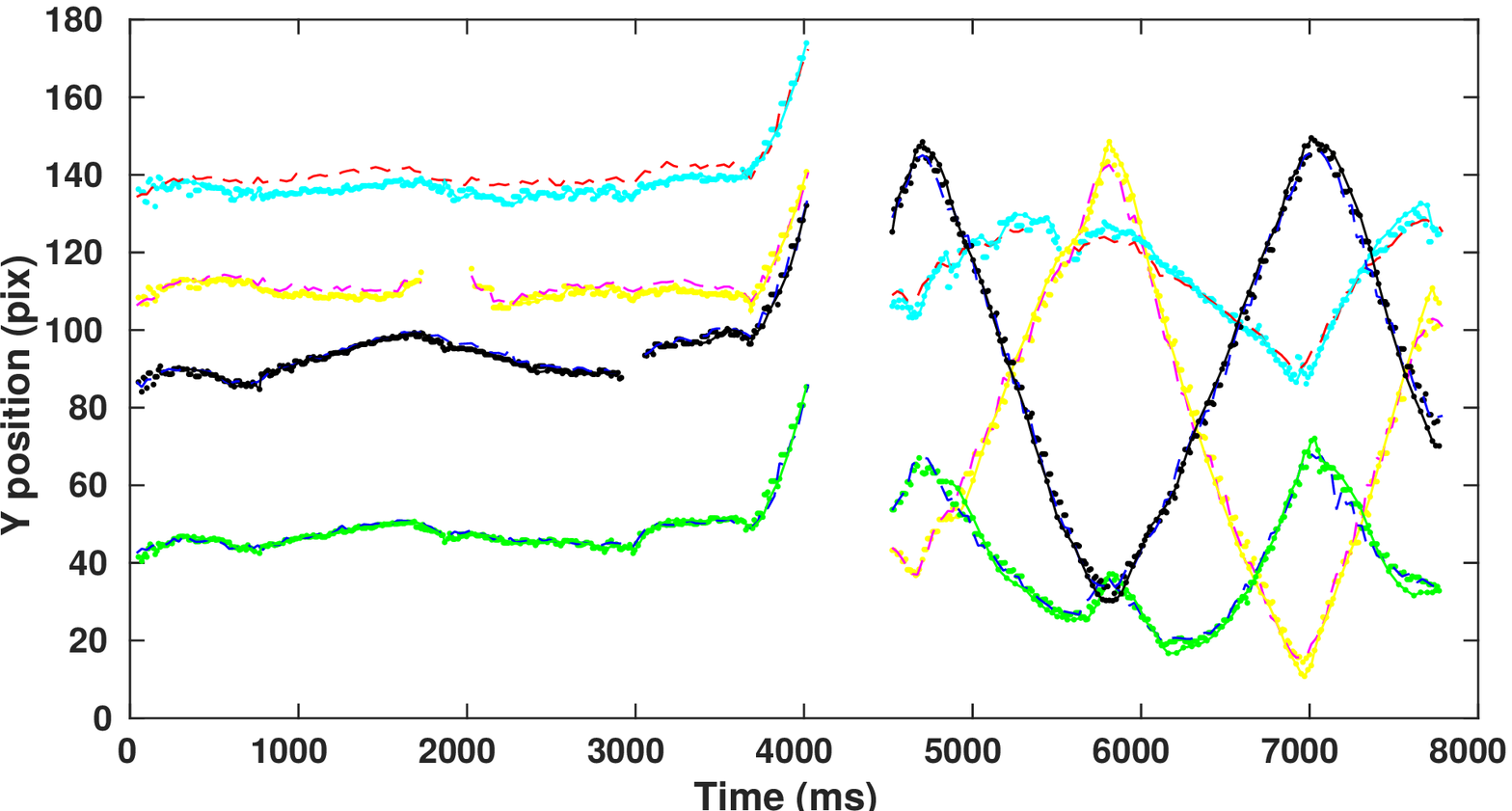}
\end{minipage}
%\hspace{0.05cm}
\begin{minipage}[b]{0.1\textwidth}
	\centering
 	\includegraphics[width=\textwidth, height=2.5cm]{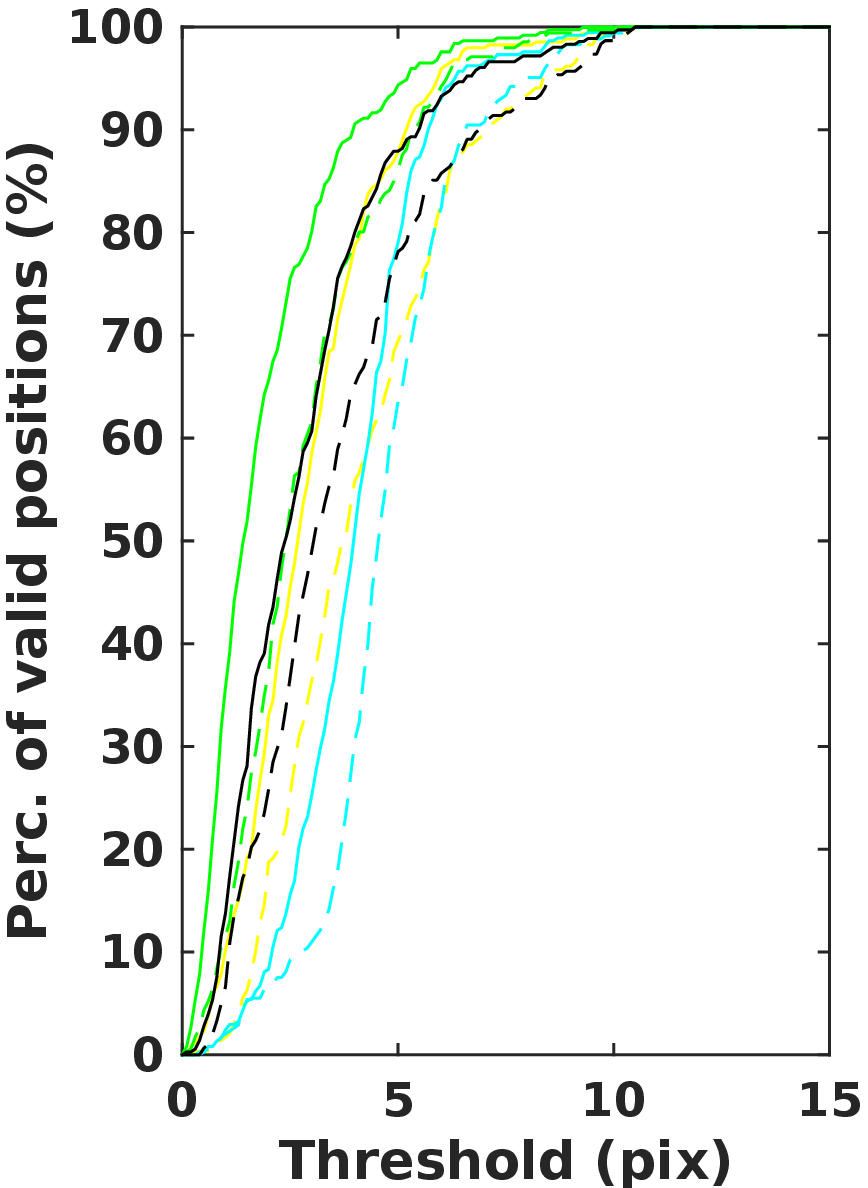}
\end{minipage}
\hspace{0.15cm}
\begin{minipage}[b]{0.075\textwidth}
	\centering
 	\includegraphics[width=\textwidth, height=1.75cm]{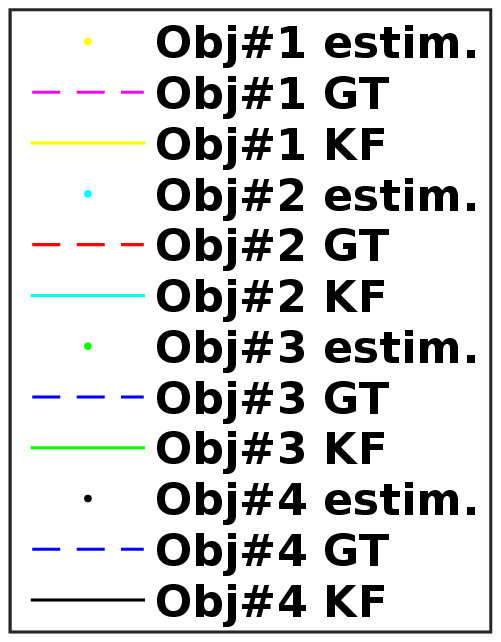}
\end{minipage}
\hspace{0.95cm}

\vspace{0.01cm}
\begin{minipage}[b]{0.32\textwidth}
	\centering
 	\footnotesize{(b)}
\end{minipage}
%\hspace{0.05cm}
\begin{minipage}[b]{0.32\textwidth}
	\centering
	\footnotesize{(e)}
\end{minipage}
%\hspace{0.05cm}
\begin{minipage}[b]{0.1\textwidth}
	\centering
 	\footnotesize{(h)}
\end{minipage}
\hspace{2.65cm}

\vspace{0.01cm}
\begin{minipage}[b]{0.32\textwidth}
	\centering
 	\includegraphics[width=\textwidth, height=2.5cm]{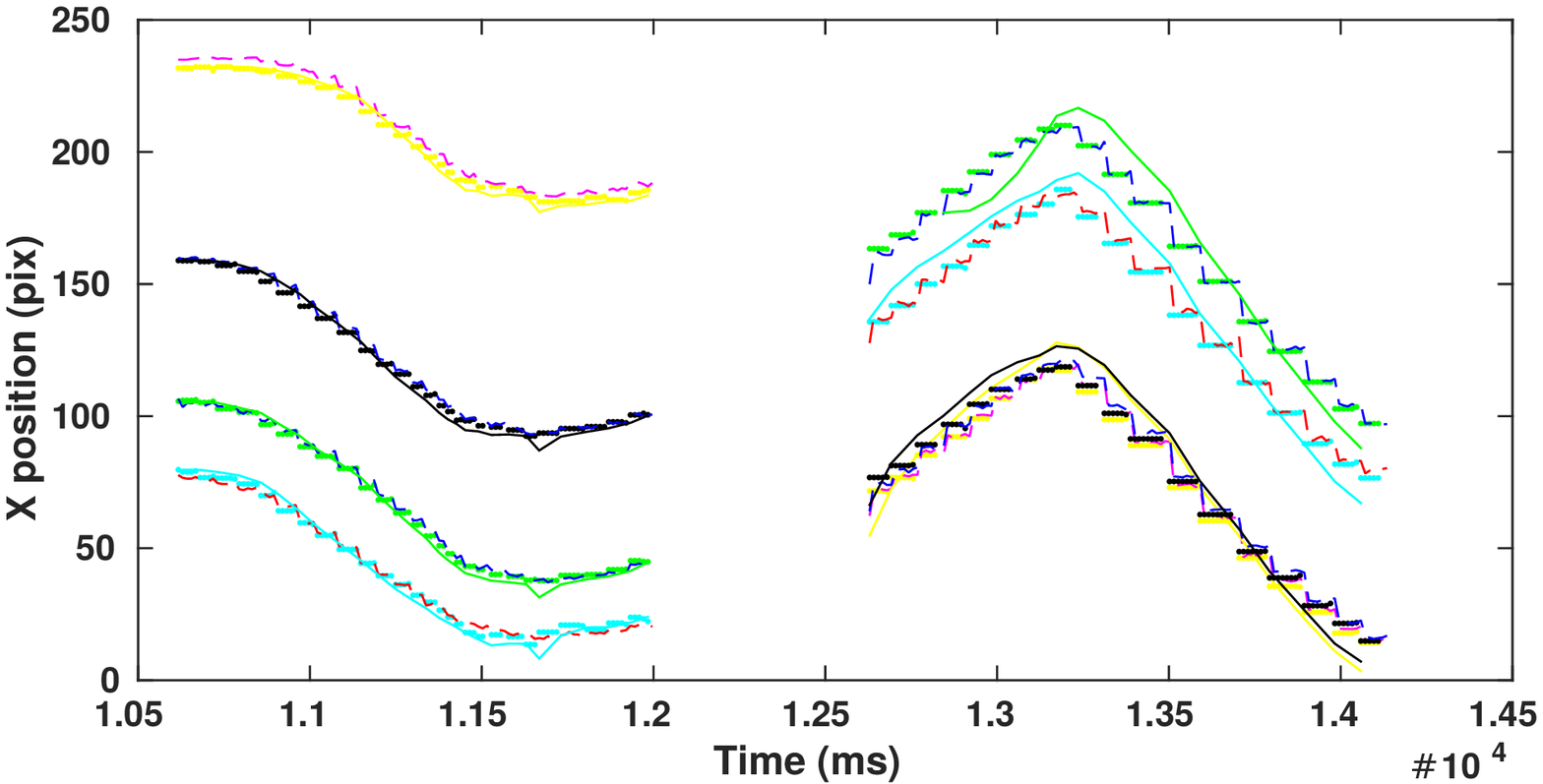}
\end{minipage}
%\hspace{0.05cm}
\begin{minipage}[b]{0.32\textwidth}
	\centering
 	\includegraphics[width=\textwidth, height=2.5cm]{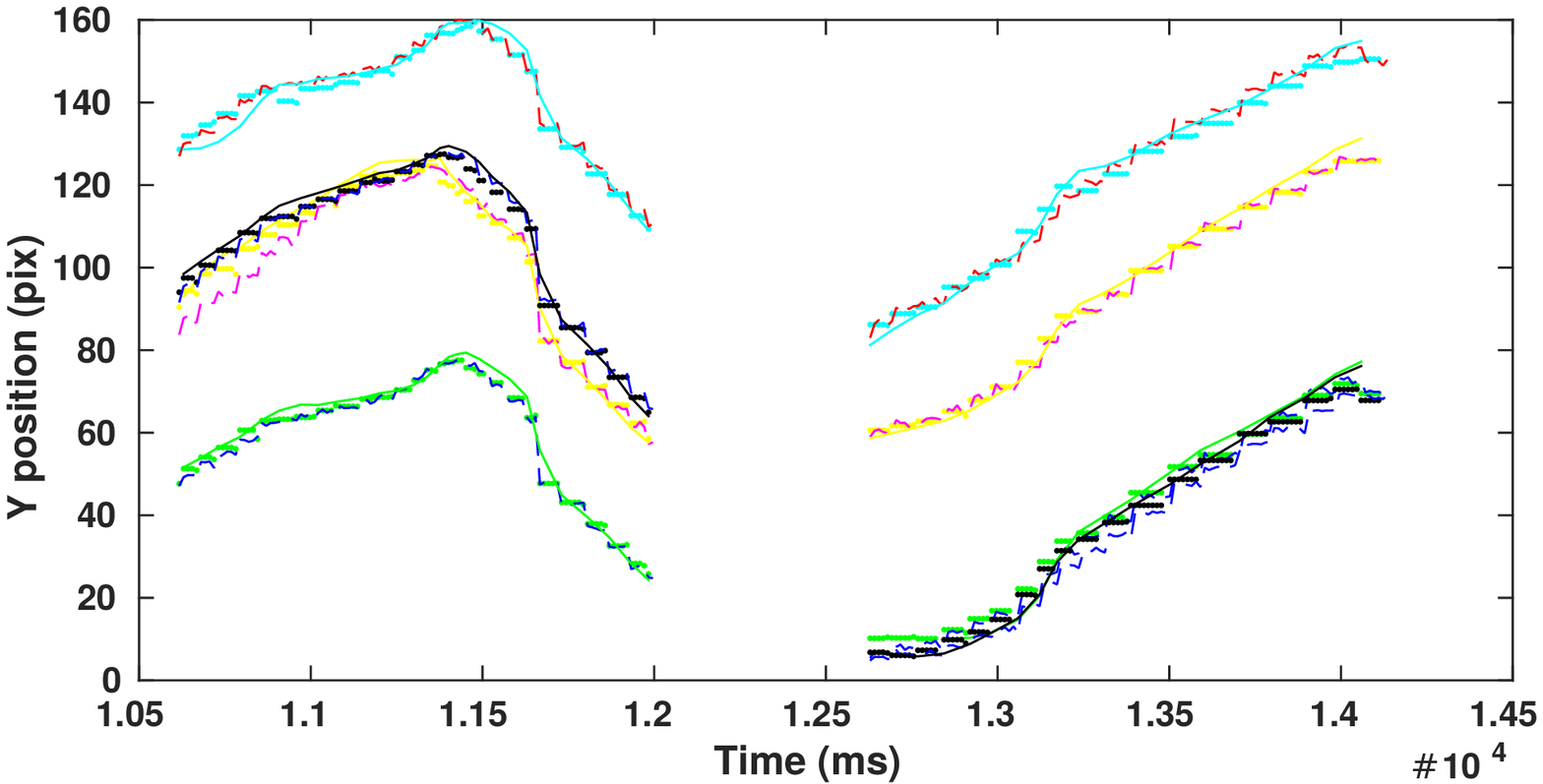}
\end{minipage}
%\hspace{0.05cm}
\begin{minipage}[b]{0.1\textwidth}
	\centering
 	\includegraphics[width=\textwidth, height=2.5cm]{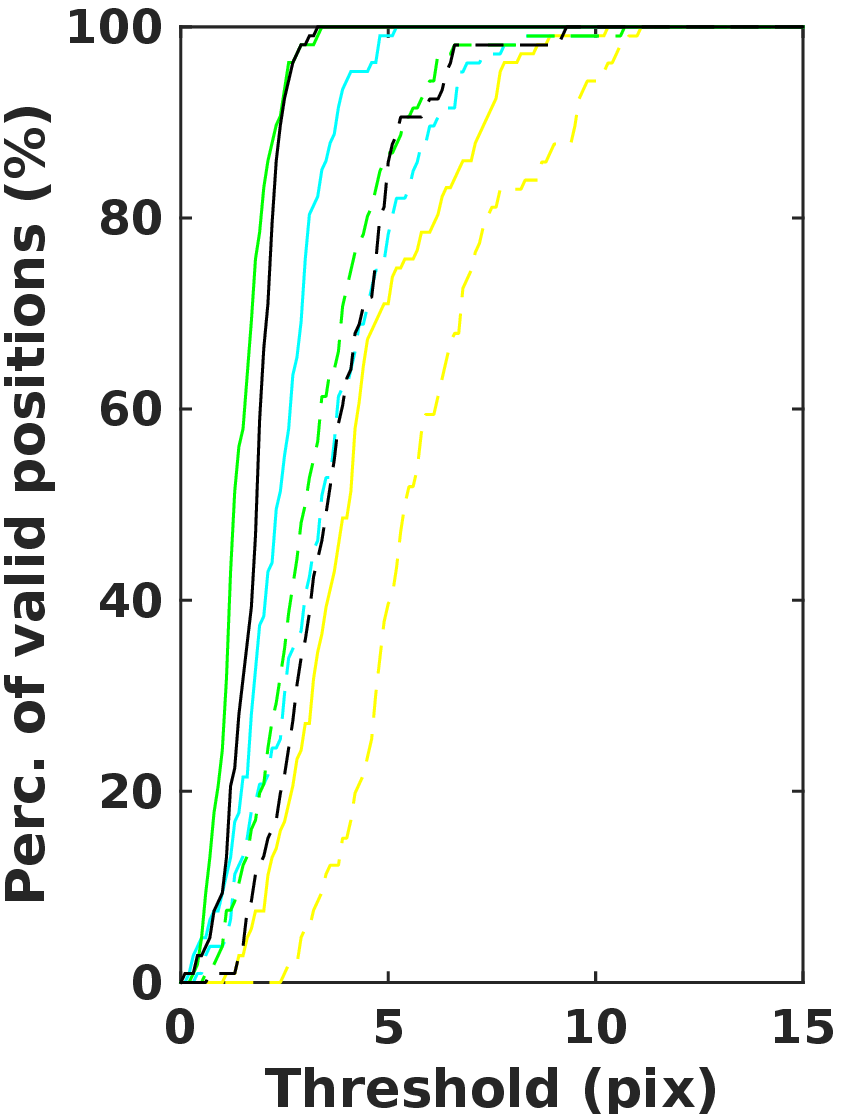}
\end{minipage}
%\hspace{0.05cm}
\begin{minipage}[b]{0.125\textwidth}
	\centering
 	\includegraphics[width=\textwidth, height=2.5cm]{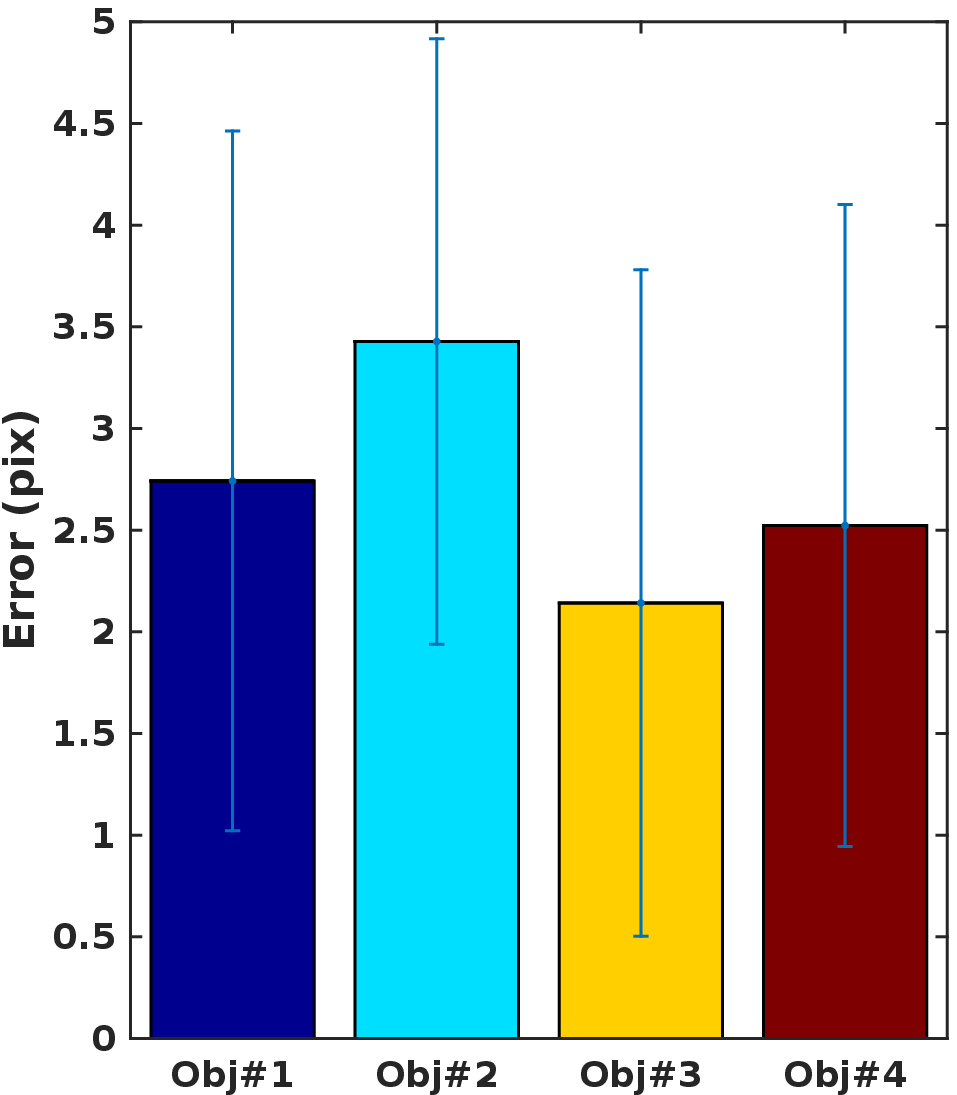}
\end{minipage}
\hspace{0.33cm}

\vspace{0.01cm}
\begin{minipage}[b]{0.32\textwidth}
	\centering
 	\footnotesize{(c)}
\end{minipage}
%\hspace{0.05cm}
\begin{minipage}[b]{0.32\textwidth}
	\centering
	\footnotesize{(f)}
\end{minipage}
%\hspace{0.05cm}
\begin{minipage}[b]{0.1\textwidth}
	\centering
 	\footnotesize{(i)}
\end{minipage}
%\hspace{0.05cm}
\begin{minipage}[b]{0.125\textwidth}
	\centering
 	\footnotesize{(j)}
\end{minipage}
\hspace{0.33cm}
\end{center}
\vspace{-3mm}
\caption{Tracking accuracy results for three sequences for the X (a,b,c) and Y (d,e,f) positions of the cluster centers. Each color encodes a different object, dotted lines represent the tracked position, solid lines the tracked position smoothed via the Kalman filtering, and dashed lines the ground-truth position (GT). (g),(h), and (i) plot the average percentage of correctly tracked positions over time for the same sequences and objects. (j) shows the average error (in pixels) and standard deviation for all sequences, for each tracked object.}
\label{fig:tracking}
\end{figure*}

Fig.~\ref{fig:performance}d shows the number of detected clusters per second with our event-based mean-shift method and the evolution for different speeds. We compute the detected clusters per second as the total count of detections along 20 seconds of each sequence. The dashed line on the bottom shows the number of clusters per second estimated for the frame-based alternative, assuming again a frame rate of 30 fps and the average number of clusters during the same 20 seconds (about 180 detections/s). The detection rate of the event-based processing increases with the speed and stabilizes when the speed factor reaches around 3x, although this number also depends on the original speed of the sequence in each case (speed factor = 1). The results for the detections here are the same at lower speeds although for a speed factor greater than 2.5 the accuracy diminishes. Let us also remark that for each detection, the frame based approach needs about 1/30~s to start detecting clusters while in our case, this latency is a few microseconds.

\subsection{Multi-target tracking evaluation}
The multi-target tracking application using the mean-shift partitioned clusters exploits the high-temporal resolution demonstrating the potential of event-based sensors for robust tracking. The evaluation of the accuracy focuses on the distance between the positions for the manually-labeled ground-truth and the tracked position of the center of masses of the cluster. We also included a posterior Kalman filtering that provides a corrected smoother result for the trajectory. Fig.~\ref{fig:tracking}a, b, c and Fig.~\ref{fig:tracking}d, e, f show respectively the $x$ and $y$ positions for two chunks of three different sequences, and four different objects encoded in different colors. Up to 15~s of sequence are shown, and the gap in the middle corresponds to the separation between the first and second chunk within the same sequence. Also, there are some parts missing (e.g. Object 1 in Fig.~\ref{fig:tracking}a and Fig.~\ref{fig:tracking}d) when the object leaves the field of view of the sensor. Object centers are successfully tracked along time although the geometric center of the labeled ground-truth object sometimes does not correspond exactly to the center of masses of the cluster. Let us explain this with an example: in Fig.~\ref{fig:bandwidth}b, the geometry of most objects is pretty well defined for the current motion, except for the rectangle and the hexagon. The rectangle (in dark blue) triggers more events in the bottom-left side that is wider because of the sensitivity of the sensor and the velocity of the motion, slightly shifting the center of masses to that side. In the case of the hexagon, some sides are lost because the direction of the motion is perpendicular to their local gradient. Locally, no change in the intensity is happening since all pixels on that side have the same intensity and thus, no events are triggered. Consequently, the location of the center of masses is affected, being more harmful with non-symmetric or irregular shapes. Despite this, the average error of the center position for each object along the sequences does not go above 2.5 pixels in the worst case (see Fig.~\ref{fig:tracking}j). Additionally, we also applied a final Kalman filtering stage to correct our predictions based on the previous history and the system model. Smoother trajectories are obtained for all sequences except for the X position of the second chunk in Fig.~\ref{fig:tracking}c. In this sequence, the image motions are due to zoom-in and out sensor motion and this affects not only the estimation but also the available data for the manual labeling. Nevertheless, we achieve lower error for the Y position in the first part of the sequence yielding to a low average error.

\begin{figure*}[tp]
\begin{center}
\begin{minipage}[b]{0.13\textwidth}
	\centering
 	\includegraphics[width=\textwidth]{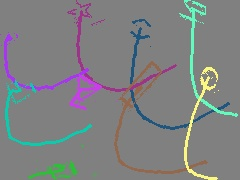}
\end{minipage}
\begin{minipage}[b]{0.13\textwidth}
	\centering
 	\includegraphics[width=\textwidth]{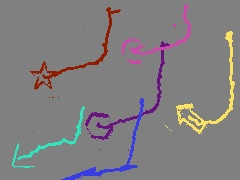}
\end{minipage}
\begin{minipage}[b]{0.13\textwidth}
	\centering
 	\includegraphics[width=\textwidth]{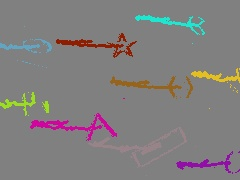}
\end{minipage}
\begin{minipage}[b]{0.13\textwidth}
	\centering
 	\includegraphics[width=\textwidth]{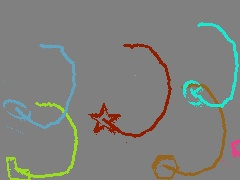}
\end{minipage}
\begin{minipage}[b]{0.1\textwidth}
	\centering
 	\includegraphics[width=\textwidth]{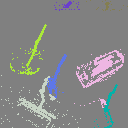}
\end{minipage}
\begin{minipage}[b]{0.1\textwidth}
	\centering
 	\includegraphics[width=\textwidth]{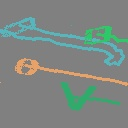}
\end{minipage}
\begin{minipage}[b]{0.1\textwidth}
	\centering
 	\includegraphics[width=\textwidth]{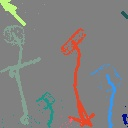}
\end{minipage}
\begin{minipage}[b]{0.1\textwidth}
	\centering
 	\includegraphics[width=\textwidth]{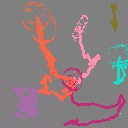}
\end{minipage}
\vspace{0.00cm}
\begin{minipage}[b]{0.13\textwidth}
	\centering
 	\includegraphics[width=\textwidth]{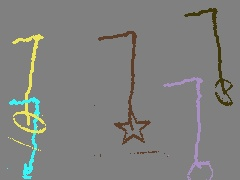}
\end{minipage}
\begin{minipage}[b]{0.13\textwidth}
	\centering
 	\includegraphics[width=\textwidth]{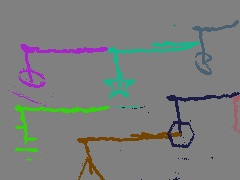}
\end{minipage}
\begin{minipage}[b]{0.13\textwidth}
	\centering
 	\includegraphics[width=\textwidth]{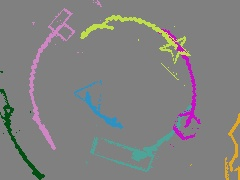}
\end{minipage}
\begin{minipage}[b]{0.13\textwidth}
	\centering
 	\includegraphics[width=\textwidth]{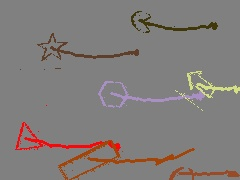}
\end{minipage}
\begin{minipage}[b]{0.1\textwidth}
	\centering
 	\includegraphics[width=\textwidth]{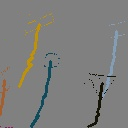}
\end{minipage}
\begin{minipage}[b]{0.1\textwidth}
	\centering
 	\includegraphics[width=\textwidth]{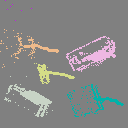}
\end{minipage}
\begin{minipage}[b]{0.1\textwidth}
	\centering
 	\includegraphics[width=\textwidth]{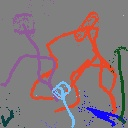}
\end{minipage}
\begin{minipage}[b]{0.1\textwidth}
	\centering
 	\includegraphics[width=\textwidth]{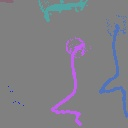}
\end{minipage}
\end{center}
\vspace{-3mm}
\caption{Examples including clustering and tracking for the sequences with patterns or the Baxter manipulation scenario. Colors encode the cluster and the solid tails show the trajectories for the last second. Accurately calculating cluster centers is harder when dealing with larger or textured objects. These images are generated plotting packets of 1500 events.}
\label{fig:examples}
\end{figure*}

We also estimated the percentage of valid tracked positions in Fig.~\ref{fig:tracking}g, h, i, for the three sequences. This metric shows the percentage of the count of estimates that are computed with an error lower than the threshold $\tau$ such that $||\hat{\textbf{p}} - \textbf{p}|| \leq \tau  $, where $\hat{\textbf{p}}$ and $\textbf{p}$ are the estimated tracked position and the ground-truth, and $||.||$ is the L-2 norm. Again, each color shows the object, and solid lines represent the estimated positions while the dashed lines are used for the positions after the Kalman filtering. In average, using a threshold $\tau$ of 2.5 pixels approximately 80\% of the positions along time for all sequences are computed correctly. 

Regarding the performance, Fig.~\ref{fig:performance}c shows very similar values for the mean-shift clustering only and the mean-shift followed by tracking, with a slight difference only for 2x speeds. Thus, the tracking and Kalman filtering does not represent a significant burden for the overall computation.

%-->In kmeans we are selecting the value for the correct number of clusters in each case to compare

%Fig. showing the setup (prob. it should be part of another fig)
\section{CONCLUSIONS}
We have presented an event-based mean-shift clustering method that, to the best of our knowledge is the first real-time event-based clustering method, and shown its application to a tracking task. Although some methods have presented solutions for tracking as mentioned in the introduction, none of them are available to be compared (we are making all our code and datasets available). While in previous attempts synchronous frame-based methods have been adapted to the asynchronous event-based framework, in our work no integration or accumulation of events is performed. For every event that inputs our pipeline, one extended event is produced with extra information that contains the label of the cluster this event belongs to.

%We present results that demonstrate that our method is invariant to the cluster shapes and to different speeds: similar accuracy results are obtained for the same trajectories executed by the Baxter robot at different speeds. Regarding tracking, our clustering information is provided with a very high temporal resolution taking advantage of the sensor asynchronous nature. This nature also helps considerably reducing the requirements for computational resources reaching an average reduction of 83\% mean-shift clustering computations when comparing to the conventional method. The experiments also show very accurate tracking examples for different sequences with an error of 2.5 pix, even when part of the objects are occluded or out of the scene. In the examples with the Baxter robot, the number of clusters was steady along time for different sequences in a typical environment for manipulation tasks. We consider this work as a first step for using real-time clustering for very fast computation of the center position and contours of objects to be grabbed by the Baxter arm. Additionally, we also consider using it for generating fast affordances of objects or real-time tracking of objects moving in dynamic environments.

Our results demonstrate that our method is very robust to cluster shapes and to different speeds: similar accuracy results were obtained for the same trajectories executed by the Baxter robot at different speeds. Regarding tracking, our clustering information is provided with very high temporal resolution taking advantage of the sensor asynchronous nature. This also helps considerably reducing the computational resources reaching an average reduction of 83\% compared to the conventional method. The experiments also show very accurate tracking examples for different sequences with an error of 2.5 pix, even when parts of the objects are occluded or out of the scene. In the examples with the Baxter robot, the number of clusters was steady along time for different sequences in a typical environment for robotic manipulation tasks. We consider this work as a first step for other  applications in robotic manipulation. For example after segmenting and tracking, Baxter can recognize and grab with its arm, or it can generate fast affordances of objects or real-time tracking of objects moving in dynamic environments.

\addtolength{\textheight}{-12cm}   % This command serves to balance the column lengths

%%%%%%%%%%%%%%%%%%%%%%%%%%%%%%%%%%%%%%%%%%%%%%%%%%%%%%%%%%%%%%%%%%%%%%%%%%%%%%%%
{\small
\bibliographystyle{ieee}
\bibliography{test}
}

\end{document}